\documentclass[10pt,twocolumn,letterpaper]{article}

\usepackage[pagenumbers]{iccv} 

\newcommand{\todo}[1]{{\color{red}#1}}

\definecolor{iccvblue}{rgb}{0.21,0.49,0.74}
\usepackage[pagebackref,breaklinks,colorlinks,allcolors=iccvblue]{hyperref}
\usepackage{algorithm2e}
\def\paperID{1786} 
\def\confName{ICCV}
\def\confYear{2025}

\title{StyleKeeper: Prevent Content Leakage using Negative Visual Query Guidance}

\author{
Jaeseok Jeong\thanks{Equal contribution}\hspace{0.4em}\thanks{This work was done during his internship at NAVER AI Lab.} \\
Yonsei University\\
{\tt\small jete\_jeong@yonsei.ac.kr}
\and
Junho Kim\footnotemark[1]\\
NAVER AI Lab\\
{\tt\small jhkim.ai@navercorp.com}
\and
Gayoung Lee\\
NAVER AI Lab\\
{\tt\small gayoung.lee@navercorp.com}
\and
Yunjey Choi\\
NAVER AI Lab\\
{\tt\small yunjey.choi@gmail.com}
\and
Youngjung Uh\thanks{Corresponding author.}\\
Yonsei University\\
{\tt\small yj.uh@yonsei.ac.kr}
}

\begin{document}

\newcommand{\js}[1]{\textcolor{blue}{#1}}
\newcommand{\jhkim}[1]{\textcolor{orange}{#1}}
\newcommand{\gy}[1]{\textcolor{blue}{#1}}

\newcommand{\uh}[1]{\textcolor{red}{#1}}
\newcommand{\mingi}[1]{\textcolor{brown}{#1}}
\newcommand{\uhc}[1]{{\footnotesize{\uh{{#1}}}}}
\newcommand{\uhcl}[1]{\uhc{$\leftarrow$#1}}
\newcommand{\uhcr}[1]{\uhc{#1$\rightarrow$}}
\newcommand{\needcite}[1]{\textcolor{red}{[needcite #1]}}
\newcommand{\Sref}[1]{Section \ref{#1}}

\newcommand{\fref}[1]{Figure~\ref{#1}}
\newcommand{\eref}[1]{Eq.~(\ref{#1})}
\newcommand{\tref}[1]{Table~\ref{#1}}
\newcommand{\sref}[1]{~\Sref{#1}}
\newcommand{\aref}[1]{Appendix \ref{#1}}
\newcommand{\dref}[1]{\ref{#1}}

\newcommand{\rfref}[1]{Fig R~\ref{#1}}

\def\injectingselfattention{injecting self-attention}
\def\Injectingselfattention{Injecting self-attention}

\def\swappingselfattention{swapping self-attention}
\def\Swappingselfattention{Swapping self-attention}
\def\negativequeryguidance{negative visual query guidance}
\def\Negativequeryguidance{Negative visual query guidance}
\def\NegativeQueryGuidance{Negative Visual Query Guidance}
\def\NVQG{NVQG}

\def\method{StyleKeeper}
\def\Method{StyleKeeper}

\def\crossstyleattention{cross style-attention}
\def\Crossstyleattention{Cross style-attention}
\def\CrossStyleAttention{Cross Style-Attention}

\def\sharedselfattention{shared self-attention}
\def\Sharedselfattention{Shared self-attention}

\def\ours{StyleKeeper}
\def\Ours{StyleKeeper}

\definecolor{darkgreen}{rgb}{0.032, 0.6392, 0.2039}
\newcommand{\cmark}{{\textcolor{darkgreen}{\ding{51}}}}%
\newcommand{\xmark}{{\textcolor{red}{\ding{55}}}}%
\maketitle
\begin{abstract}

In the domain of text-to-image generation, diffusion models have emerged as powerful tools. Recently, studies on visual prompting, where images are used as prompts, have enabled more precise control over style and content. However, existing methods often suffer from content leakage, where undesired elements of the visual style prompt are transferred along with the intended style. To address this issue, we 1) extend classifier-free guidance (CFG) to utilize \swappingselfattention{} and propose 2) \negativequeryguidance{} (\NVQG) to reduce the transfer of unwanted contents. \NVQG{} employs negative score by intentionally simulating content leakage scenarios that swap queries instead of key and values of self-attention layers from visual style prompts. This simple yet effective method significantly reduces content leakage. Furthermore, we provide careful solutions for using a real image as visual style prompts.
Through extensive evaluation across various styles and text prompts, our method demonstrates superiority over existing approaches, reflecting the style of the references, and ensuring that resulting images match the text prompts. Our code is available \href{https://github.com/naver-ai/StyleKeeper}{here}.


\end{abstract}    
\section{Introduction}
\label{introduction}

\begin{figure}[t]
    
        \centering
        \includegraphics[width=\linewidth]{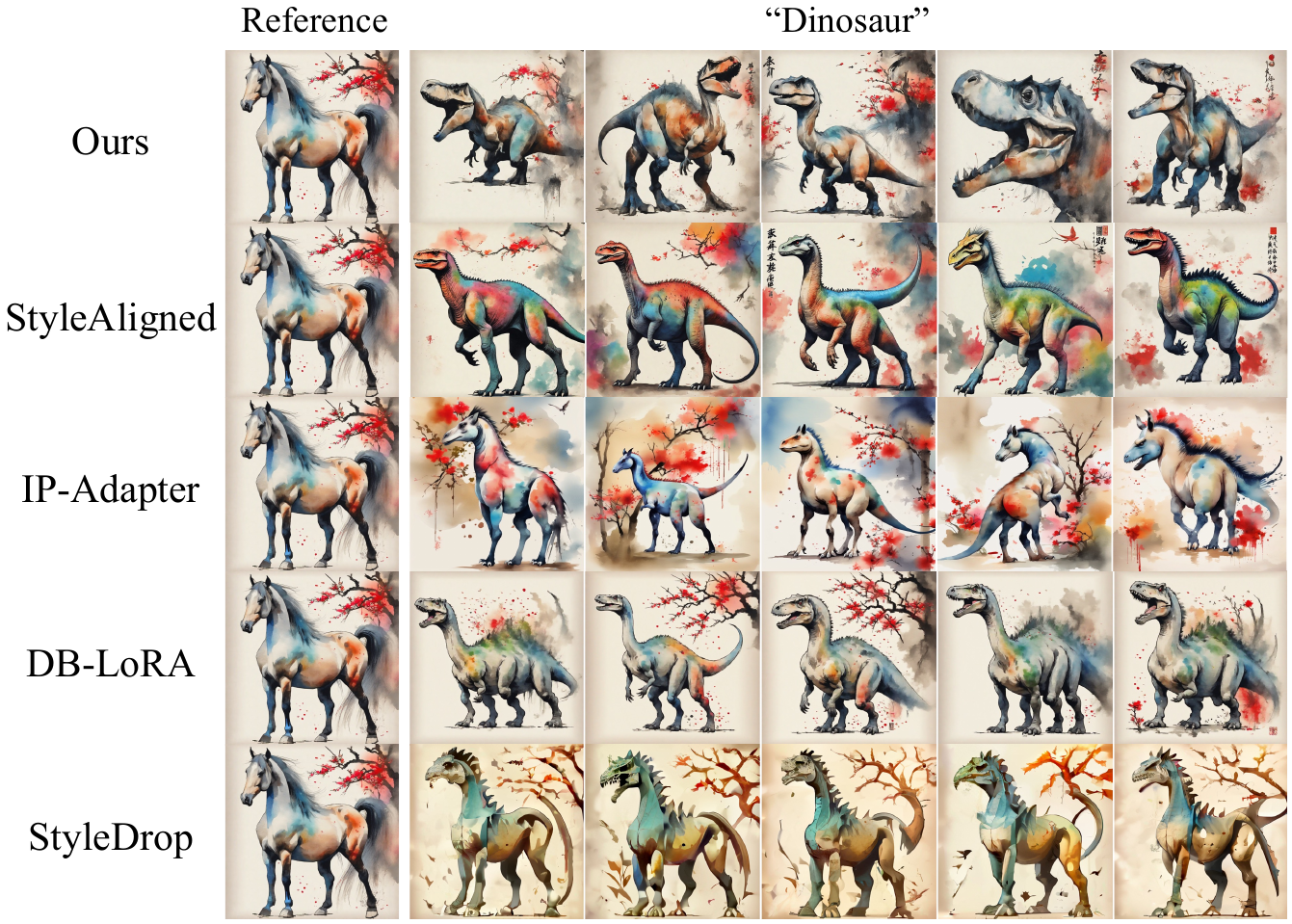}
        \caption{\textbf{Qualitative comparison with the same style.} Competitors face challenges in generating images with diverse layouts and compositions, i.e., content leakage from the reference.}
        \label{fig:diversity_vs_competitors}
\end{figure}

Text-to-image diffusion models (T2I DMs) excel at synthesizing images that correspond to given text prompts \citep{rombach2022high,ramesh2021zero}. However, relying solely on text prompts may not allow precise control over the desired output. Even with highly detailed text prompts, controlling the exact style of the resulting images remains a challenge because of the modality gap between text and image. For example, from ``a red car", different people imagine different variants of red. Instead, ``A dinosaur" drawn in the style of the reference image in \fref{fig:diversity_vs_competitors} provides a more concrete specification of the desired style.

However, using reference images as visual style prompts comes with a fundamental challenge: a dilemma between better reflecting style and preventing content leakage.
\fref{fig:diversity_vs_competitors} shows that the pose of the reference horse appears in the results of previous methods.
The content leakage may include poses, specific shapes, layouts or entire objects.
Furthermore, content leakage reduces diversity and text alignments as shown in \fref{fig:diversity_vs_competitors}.

Approaches that involve training the model, such as fine-tuning a diffusion model on a set of themed images \citep{ruiz2023dreambooth, kumari2023multi}, learning new text embeddings \citep{gal2022image, han2023highly, avrahami2023bas}, or adapting cross-attention modules to incorporate image features \citep{ye2023ip, wang2023styleadapter, xing2024csgo, qi2024deadiff}, come with significant drawbacks. These methods not only incur high computational costs, but also face an inherent trade-off between style and content. Achieving a balance where the model faithfully captures the desired style without unintentionally preserving content remains a challenge.

\begin{figure*}[t]
    \centering
    \includegraphics[width=0.9\textwidth]{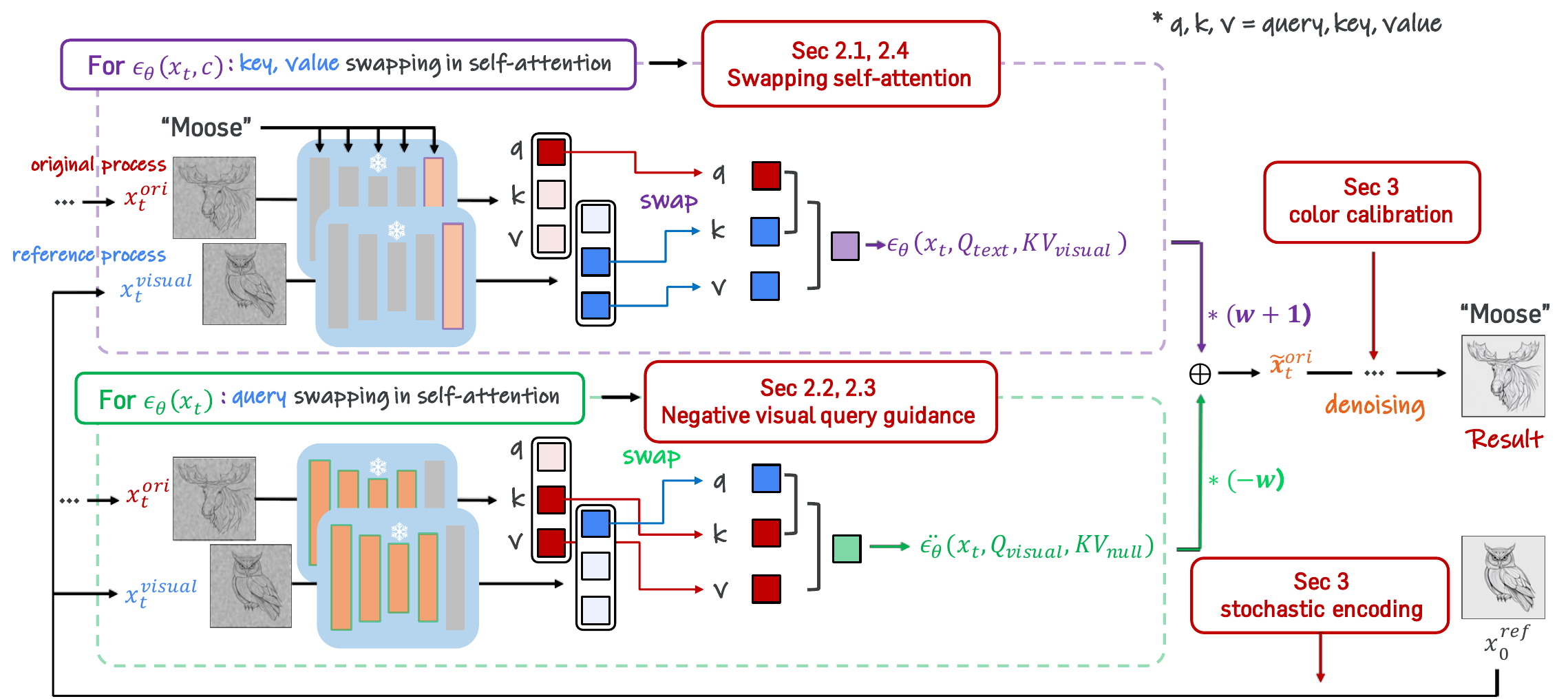}
    \caption{\textbf{Overveiw of \method}.
    Our proposed method includes 4 key components, highlighted in red boxes. First, stochastic encoding (Section \ref{sec:real_image_as_reference}) converts reference images into suitable latents for the visual style prompting task. Second, swapping self-attention (Section \ref{sec:swapping_selfattention}, \ref{sec:choosing_blocks}) ensures the reference image’s style is accurately reflected. Third, \negativequeryguidance{} (Section \ref{sec:negativequeryguidance}) reduces content leakage from the reference image, allowing the desired text content (e.g., ``Moose'') to be better represented. Lastly, color calibration (Section \ref{sec:real_image_as_reference}) minimizes errors during the denoising process, helping to produce a cleaner final image.}
    \label{fig:overview}
\end{figure*}

To address this, recent papers \citep{hertz2023style, chung2024style, alaluf2024cross} have proposed training-free methods by modifying feature swapping mechanisms within the self-attention layers. Specifically, they replace the key and value features in the original process with those extracted from the visual style prompt. This allows the query to preserve content while the key-value pair carries the style, enabling promising style transfer performance.

However, these methods struggle to fully eliminate content leakage \citep{sohn2023styledrop}. Additionally, \citet{hertz2023style} does not support real images as reference inputs, while \citet{chung2024style} and \citet{alaluf2024cross} primarily focus on image-to-image (I2I) tasks. When using real images as style references, they often suffer from either poor style transfer or excessive content leakage, significantly limiting their effectiveness.

In this study, we propose a method to separately control the strength of style and content from visual prompts such that the results align with the text prompts.
We design a novel variation of classifier-free guidance (CFG)~\citep{ho2022classifier}, emulating images with undesirable content leakage as a negative guidance. To address content leakage, we introduce negative visual query guidance, ensuring a clear separation between content and style. We also show that \textit{stochastic encoding} achieves better style reflection with less artifacts and computation, and incorporate \textit{color calibration} to precisely match the final output to the reference image's color statistics leading to better style reflection.


We analyze where to apply swapping self-attention, identifying the optimal layers for balancing style transfer and content fidelity. Additionally, our method can effectively remove content that is difficult to eliminate with key and value swapping alone, working successfully even in cases with significant structural gaps between the style image and content text, as shown in \fref{fig:diversity_vs_competitors} (e.g., a complex scene of ``a horse with the vibrant colors of a Chinese ink painting'' and a single object ``Dinosaur''). Our method is well compatible with ControlNet for I2I style transfer, further enhancing its flexibility. Qualitative and quantitative evaluations show our method outperforms state-of-the-art approaches, providing precise control over content and style without content leakage. Our approach is both robust and efficient, ideal for visual style prompting tasks.\footnote{Our code will be released for reproducibility.}

\section{Visual style prompting}
We propose \ours{} which receives a text prompt and a visual style prompt to generate new images without content leakage. The results contain the content and style specified by the text prompt and the visual style prompt, respectively, with variations due to initial noises. The overview of our method is illustrated in \fref{fig:overview}.
First, we explain the swapping self-attention in the aspect of style transfer literature.
\Ours{} consists of classifier-free guidance (CFG) with swapping self-attention, \negativequeryguidance{} (NVQG), optimal layer choice, stochastic encoding of real visual style prompts, and generalization to ControlNet for real content images.
We explain the first three components in text-to-image (T2I) scenario with generated visual style prompt. Then we proceed to T2I with real visual style prompt and image-to-image (I2I) scenario.

\subsection{Swapping self-attention in style transfer literature}
Modern diffusion models consist of a number of self-attention and cross-attention blocks \citep{vaswani2017attention}. Both of them employ the attention mechanism, which first obtains an attention map using similarity between query features $Q$ and key features $K$, then aggregates value features $V$ using the attention map as weights: $\text{Attention}(Q, K, V) = \text{Softmax}(\frac{QK^T}{\sqrt{d}})V$.
Opposed to the cross-attention layer, self-attention layer receives key and values coming from the main denoising process, which has spatial dimensions with more freedom to represent spatially varying visual elements. As our goal is to reflect style elements from a reference image that are not easily represented by textual description, we opt to borrow key and values of self-attention layers in the reference process to the original process, namely swapping self-attention (\fref{fig:overview}).
In addition, swapping self-attention has a strong connection with style transfer literature \citep{sheng2018avatar,park2019arbitrary,yao2019attention,liu2021adaattn, deng2022stytr2}. where the attention mechanism reassembles visual features of a style image (key, value) on a content image (query).  However, methods \cite{hertz2023style, cao_2023_masactrl} utilizing self-attention operation often exhibit artifacts and content leakage since style elements are entangled with content elements in the visual feature.

\subsection{CFG sampling with \swappingselfattention{} for T2I}
\label{sec:swapping_selfattention}



Previous works \cite{alaluf2024cross, chung2024style} have not taken CFG into account while manipulating features. We propose CFG combined with \swappingselfattention{} to reflect a visual style prompt in the T2I results without artifacts.
CFG \citep{ho2022classifier} is a key technique that steers generated images toward specified text prompts. Given a score $\epsilon_\theta(x_t,c)$ conditioned on $c$ and an unconditional score $\epsilon_\theta(x_t,\emptyset{})$, the CFG score is defined as:
\begin{equation}
    \tilde{\epsilon}_\theta = (1+w)\epsilon_\theta(x_t,c) - w\epsilon_\theta(x_t,\emptyset{}), 
\end{equation}
\footnote{We omit the diffusion timestep $t$ in the arguments and abuse $x_t$ instead of $z_t$ from latent diffusion model.}
where $w$ controls guidance strength. CFG with $w>1$ enhances image quality and text alignment, but excessive $w$ may lead to mode collapse \citep{chung2024cfg++}.
Notably, CFG has not been explored in the context of reflecting denoising process with modified features.

Assuming a given condition $c$ has hidden content $h^\text{content}$ and style $h^\text{style}$, we model $p_\theta(x_0 | h^\text{content}_\text{text}, h^\text{style}_\text{visual})$ using:

\begin{itemize}
    \item The original T2I score: $\epsilon_\theta(x_t, c_\text{text})$, leading to $x_0^\text{ori}\sim p_\theta(x_0 | c_\text{text}) = p_\theta(x_0 | h^\text{content}_\text{text}, h^\text{style}_\text{text})$.
    \item The reference score: $\epsilon_\theta(x_t, c_\text{visual})$, leading to $x_0^\text{visual}\sim p_\theta(x_0 | c_\text{visual}) = p_\theta(x_0 | h^\text{content}_\text{visual}, h^\text{style}_\text{visual})$.
\end{itemize}
To ensure proper content and style control, we design the CFG score as:
\begin{equation}
    \label{eq:cfg}
    \begin{aligned}
        \tilde{\epsilon}_\theta(x_t,h^\text{content}_\text{text},h^\text{style}_\text{visual}) 
        & = (1+w) \\ & \quad \ddot{\epsilon}_\theta(x_t,Q_\text{text},{KV}_\text{visual}) \\
        & \quad - w\epsilon_\theta(x_t,\emptyset{}).
    \end{aligned}
\end{equation}

where $\ddot{\epsilon}_\theta(x_t,Q_\text{text},{KV}_\text{visual}) $ represents a KV-injected denoising score, indicating feature manipulation rather than direct conditioning.

For given two denoising processes, one as original and another as a reference, borrowing the key-value in self-attention from the reference to the original process, i.e., key-value (KV) injection, tends to produces results with the content from the original process and the style from the reference process with limited control \citep{alaluf2024cross,chung2024style,xu2023inversion}. 

We define the KV-injected score by
$\ddot{\epsilon}_\theta(x_t,Q_\text{text},{KV}_\text{visual})$
where $Q_\text{text}$ and ${KV}_\text{visual}$ denote the query from the original score\footnote{The original score and its query within are recursively altered by KV injection along the denoising process.} $\epsilon_\theta(x_t,c_\text{text})$ and the key, value from the reference score $\epsilon_\theta(x_t,c_\text{visual})$. Then KV-injected-Attention becomes:
\small
\begin{equation}
\label{eq:kvinjection}
    \begin{aligned}
        \text{Attention}(Q_\text{text}, K_\text{visual}, V_\text{visual}) &= \text{Softmax}\left(\frac{Q_\text{text} K_\text{visual}^\intercal}{\sqrt{d}}\right)V_\text{visual}.
    \end{aligned}
\end{equation}
\normalsize
We omit the layer index for brevity.

Naive denoising process with $\ddot{\epsilon}_\theta(x_t,Q_\text{text},{KV}_\text{visual})$ provides limited control in generating images with content and style specified by a text $c_\text{text}$ and a visual style prompt, respectively. Moving forward, our CFG with \swappingselfattention{} in \eref{eq:cfg} enjoys higher image quality and more accurate text alignment than the naive denoising process as in the original CFG for T2I generation. The results are deferred to \sref{exp:cfgnqg}. However, content leakage still hinders the creation of diverse images, and disrupts text alignment.

\subsection{\Negativequeryguidance}
\label{sec:negativequeryguidance}
We propose \negativequeryguidance~(\NVQG) to mitigate content leakage by reducing the content $h_\text{visual}^\text{content}$ from the visual style prompt appearing in the results.
Briefly, \NVQG~negates the CFG of a score $\ddot{\epsilon}(x_t, Q_\text{visual}, {KV}_\text{text})$.

In \citet{liu2022compositional}, a complex text condition $c$ is factorized as a set of conditions $\{c_0,c_1,...\}$
and Bayes' rule induces $p_\theta(x_t|c_0,c_1,...) \propto \Pi{\frac{p_\theta(x_t|c_i) }{p_\theta(x_t)}}$. 
Then, the score of the complex text condition $c$ becomes
$\epsilon{}_{\theta}(x_t,c) = \epsilon{}_{\theta}(x_t,\emptyset{}) + \Pi{(\epsilon{}_{\theta}(x_t,c_i) - \epsilon{}_{\theta}(x_t,\emptyset{}))}$. It allows reducing a specific concept $\tilde{c}$ with composition 
by negating its guidance with scale $w_\text{neg}$:
\begin{equation}
\label{eq:negativeprompt}
\epsilon{}_{\theta}(x_t,c, \text{not }\space \tilde{c}) = \epsilon{}_{\theta}(x_t,c) - w_\text{neg}(\epsilon{}_{\theta}(x_t,\tilde{c}) - \epsilon{}_{\theta}(x_t,\emptyset{}))
\end{equation}
Although we design $\ddot{\epsilon}_\theta(x_t,Q_\text{text},{KV}_\text{visual})$ to predict the score toward $p_\theta(x_0|h_\text{text}^\text{content},h_\text{visual}^\text{style})$, $\ddot{\epsilon}(x_t,Q_\text{text},{KV}_\text{visual})$ still contain $h_\text{visual}^\text{content}$. Assuming a hidden factorization ${KV}_\text{visual}=\{{KV}_\text{visual}^\text{content},{KV}_\text{visual}^\text{style}\}$, Bayes' rule induces
\small
\begin{equation}
    \begin{aligned}
        p_{\theta}(x_t|Q_\text{text},{KV}^\text{style}_\text{visual},{KV}^\text{content}_\text{visual}) &\propto p_{\theta}(x_t) \frac{p_{\theta}(x_t|Q_\text{text},{KV}^\text{style}_\text{visual})}{p_{\theta}(x_t)} \\
        &\quad \times \frac{p_{\theta}(x_t|Q_\emptyset{},{KV}^\text{content}_\text{visual})}{p_{\theta}(x_t)}.
    \end{aligned}
\end{equation}
\normalsize

Since $\epsilon_{\theta{}}(x_t) = \epsilon_{\theta{}}(x_t,Q_\emptyset{},{KV}_\emptyset{})$, we obtain the desired conditional score of $\hat{c} = \{Q_\text{text},{KV}^\text{style}_\text{visual}\}$:
\begin{equation}
\label{eq:kvdecomposition}
    \begin{aligned}
        \epsilon_{\theta{}}(x_t,\hat{c}) \xleftarrow{} &\ w_\text{visual}(\ddot{\epsilon}_{\theta{}}(x_t,Q_\text{text},{KV}_\text{visual}) - \epsilon_{\theta{}}(x_t)) \\
        & - w_\text{content}(\ddot{\epsilon}_{\theta{}}(x_t,Q_\emptyset{},{KV}^\text{content}_\text{visual}) - \epsilon_{\theta{}}(x_t)) \\
        & + \epsilon_{\theta{}}(x_t).
    \end{aligned}
\end{equation}

where $w_\text{visual}$ and $w_\text{content}$ regulate the strength of each guidance. By leveraging query injection \cite{tumanyan2023plug, alaluf2024cross, chung2024style, xu2023inversion}, we approximate $\ddot{\epsilon}_{\theta{}}(x_t,Q_\emptyset{}, {KV}^\text{content}_\text{visual}) \approx \ddot{\epsilon}_{\theta{}}(x_t,{Q}_\text{visual}, {KV}_\emptyset{})$.
Finally, inserting \eqref{eq:kvdecomposition} into \eqref{eq:cfg}, and empirically replacing $\epsilon_{\theta{}}(x_t)$ with $\ddot{\epsilon}{\theta{}}(x_t,{Q}_\text{visual})$ results in similar performance. This allows a simplified reformulation of diffusion sampling using $w' = w_\text{visual}(w + 1)$:
Since we can highly relate the $w'\ddot{\epsilon}_{\theta{}}(x_t,{Q}_\text{visual})$ to the concept negation in \eqref{eq:negativeprompt} which guides the negation of concept with a scale $w_\text{neg}$, we named the query term as \negativequeryguidance{}.

\begin{figure}[t]
        \centering
        \includegraphics[width=\linewidth]{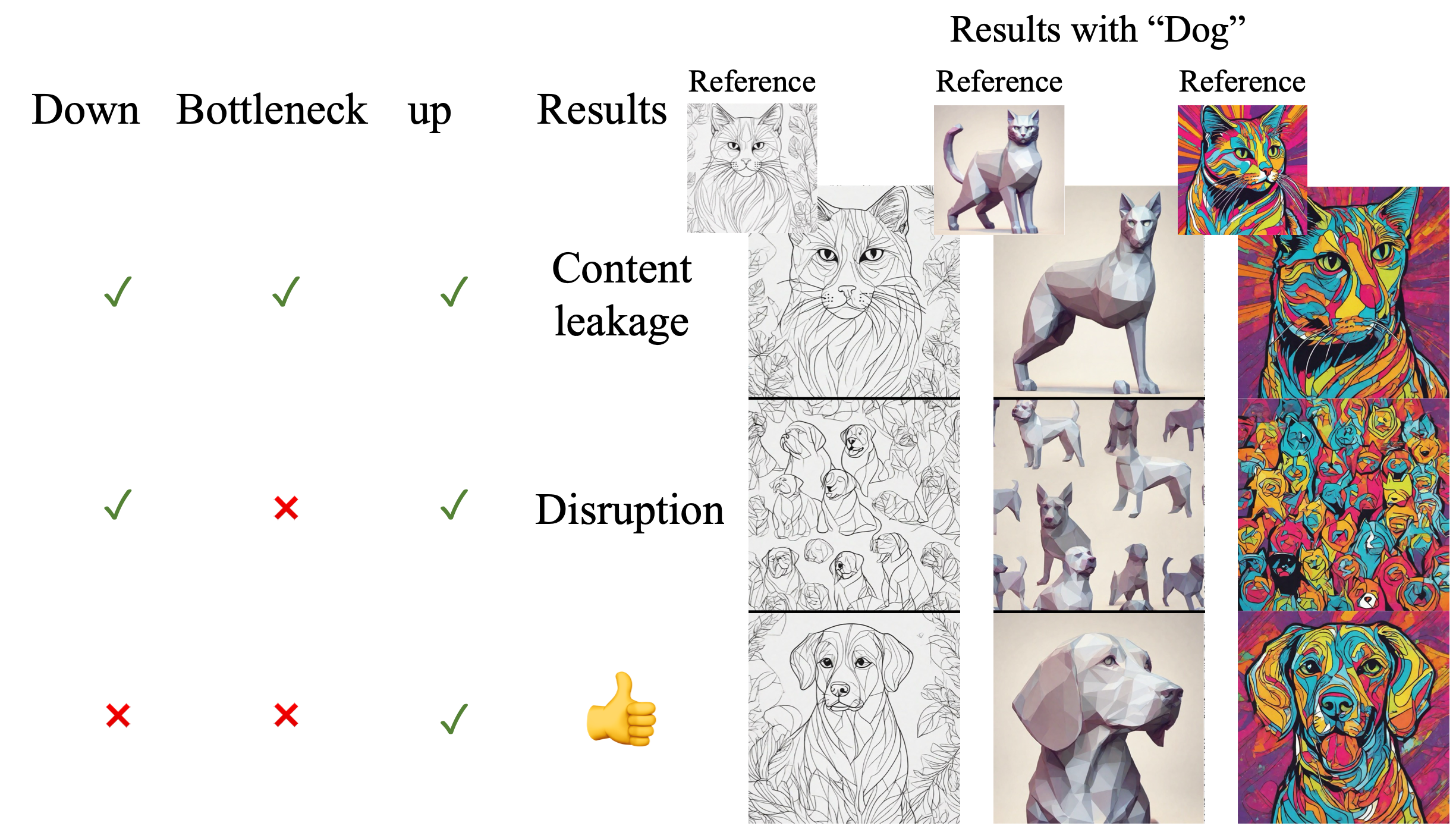}
        \caption{
        \textbf{The effect of 
        swapping self-attention across different blocks.}
        Swapping self-attention on the bottleneck and downblocks causes content leakage, resulting in cat-like images despite a dog prompt, while swapping on downblocks disrupts resulting images. We only apply swapping self-attention in the upblocks to reflect style elements effectively.
        }
        \label{fig:why_only_upblock}
\end{figure}

\subsection{Choosing blocks for swapping self-attention}
\label{sec:choosing_blocks}
In order to maximize the effect of NVQG in preventing content leakage, we explore the depth of the self-attention blocks to be swapped in the sense of granularity of visual elements.
Modern architecture of diffusion models roughly consists of three sections in a sequence: downblocks, bottleneck blocks, and upblocks. Given that the bottleneck of diffusion models contains content elements of the image~\citep{kwon2023diffusion,jeong2024training,park2023understanding}, we opt not to apply swapping self-attention to bottleneck blocks to prevent transferring contents in a reference image. \fref{fig:why_only_upblock} shows that not swapping the bottleneck blocks prevents content leaking from the reference image. However, the synthesized images show disrupted results with seriously scattered objects. Furthermore, while swapping self-attention implements the reassembling operation, simply applying to all self-attention layers exposes a content leakage problem, where the content of the reference image influences the resulting image, as shown in the first row of \fref{fig:why_only_upblock}. I.e., the results contain cats even though the prompts specify ``a dog". We conjecture that this phenomenon happens because feature maps of downblocks have unclear content layout~\citep{cao_2023_masactrl, meng2024not}, so substituting features based on this inaccurate layout causes the disrupted results. To avoid injecting unnecessary features, we choose to swap the key and value of self-attention only in upblocks. 
\begin{figure}[t]
    \centering
        \includegraphics[width=0.9\linewidth]{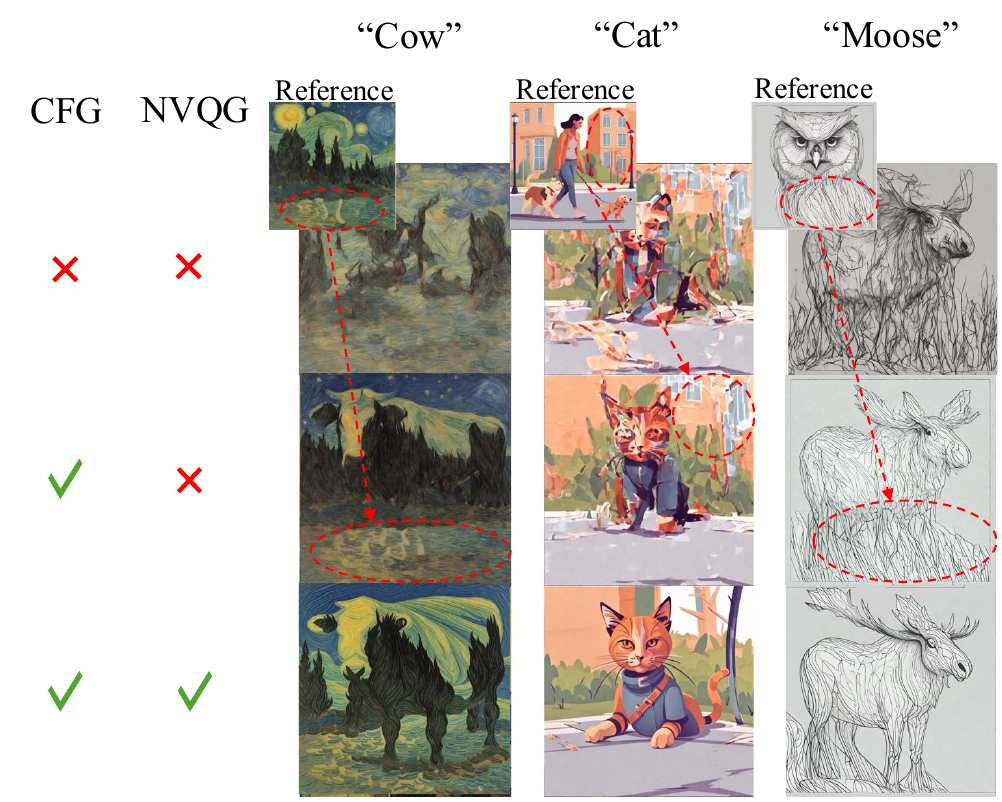}
        \caption{ 
        \textbf{The effect of CFG and the proposed \negativequeryguidance{} on image generation.} The reference images provide the style for each generated output. Without \NVQG, content leakage occurs, and the generated images fail to fully capture the intended content. In contrast, using \NVQG~ensures better alignment with both the reference style and the ``Cat'' prompt, reducing content distortion and improving quality.}
        \label{fig:cfg_nqg}
        \vskip -0.16in
\end{figure}

\begin{figure}[t]
    \centering
    \includegraphics[width=0.9\linewidth]{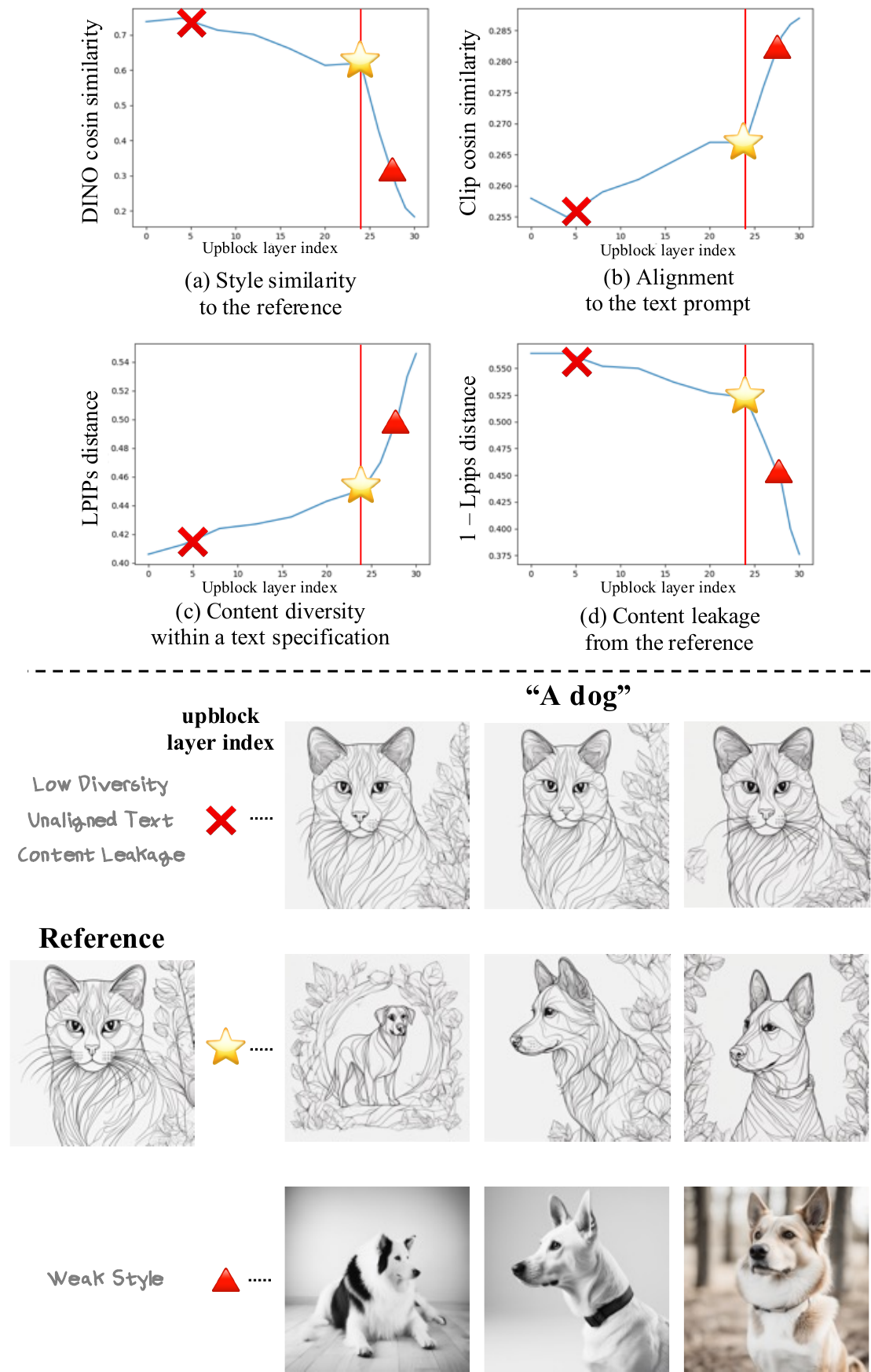}
    \caption{\textbf{Analysis on the optimal range of upblocks for swapping self-attention}.
    We find the optimal range of upblocks for a balanced trade-off between different aspects. The images on the right illustrate the visual results for different upblock layer indices, with the red cross indicating poor diversity and misalignment to the text prompt, the red triangle indicating a lack of style similarity, and the yellow star indicating the optimal results. Please refer to \sref{sec:exp_swap_sa} for details.}
    \label{fig:ablation:upblock}
\end{figure}

We note that \citet{hertz2023style} applies self-attention operation at all blocks and suffers content leakage. The last row of \fref{fig:why_only_upblock} shows the success of our strategy. Following this, we proceed to search for a specific optimal layer in the upblock in \sref{sec:exp_swap_sa}
\section{Real images as references}
\label{sec:real_image_as_reference}
So far, we have assumed a \textit{generated} visual style prompt $x_0^\text{visual}\sim p_\theta(x_0|c_\text{visual})$. Here, we allow \textit{real} visual style prompts by 1) stochastic encoding and 2) color calibration.

We propose stochastic encoding to obtain $x_t^\text{visual}\sim q(x_t|x_0^\text{visual})$ by adding a random noise on $x_0^\text{visual}$ following the forward process of DMs \citep{ho2020denoising}:
\begin{equation}
\label{eq:stochastic}
        \epsilon_t \sim \mathcal{N}(0, I),\text{ }
    x_t^\text{visual} = \sqrt{\alpha_t} \cdot x_0^\text{visual} + \sqrt{1 - \alpha_t} \cdot \epsilon_t
\end{equation}
At each timestep, we sample $\epsilon_t$ to encode $x_t^\text{visual}$.
It ensures that $x_t^{\text{visual}}$ lies on the learned trajectory and does not suffer from accumulative numerical error due to iterative process of DDIM inversion used by previous methods \citep{hertz2023style,chung2024style} as shown in \fref{afig:p-values_of_gaussian}.
Furthermore, it does not need to store the intermediate latents as opposed to DDPM inversion used by \citet{alaluf2024cross}.

Although stochastic encoding performs better than DDIM inversion, a subtle color discrepancy occurs between the resulting images and the visual style prompts. We introduce color calibration at $x_t^\text{ori}$ in the original process to match the statistics of predicted $x_0^\text{ori}$ to predicted $x_0^\text{visual}$. In \citet{gatys2015neural}, distance between channel-wise statistics is employed as a style loss for style transfer. In \citet{song2020denoising}, predicted $x_0$ $( =  \frac{x_t^\text{visual} - \sqrt{1 - \alpha_t} \cdot \epsilon_\theta(\hat{x}_t)}{\sqrt{\alpha_t}})$ allows to estimate $x_0$ with high probability at intermediate timesteps using deterministic sampling. Inheriting the advantage, we execute adain operation to match mean\&std of predicted $x_0^\text{ori}$ with those of $x_0^\text{visual}$. It allows precise color calibration rather than directly matching $x_t^\text{ori}$ to $x_t^\text{visual}$ in \citet{alaluf2024cross, chung2024style}. Furthermore, our approach differs from \citet{chung2024style} in that using predicted $x_0$ at intermediate timestep $t \in (0, T)$  other than $x_T$ by reducing cumulative sampling error after the operation. Furthermore, \citet{chung2024style} executes AdaIN at timestep T, inducing lengthy cumulative error.

We provide supportive experiments that show the effectiveness of the proposed method in \sref{sec:exp_real_image} and a detailed Algorithm in \aref{sec:algorithm}.

\section{Experiments}
In this section, we describe the effects of our proposed methods: CFG with swapping self-attention, Negative visual query guidance (\NVQG), stochastic encoding, and color calibration. For swapping self-attention, we provide a detailed analysis through experiments to determine the optimal layers for swapping. The impact of \NVQG{} is demonstrated through qualitative results. Additionally, we show why stochastic encoding outperforms DDIM inversion when inverting real images, and we highlight the benefits of color calibration through experimental results.

We also conducted both quantitative and qualitative comparisons of our method against various competitors, including StyleAligned \citep{hertz2023style}, IP-Adapter \citep{ye2023ip}, Dreambooth-LoRA \cite{ruiz2023dreambooth,lora_repo}, StyleDrop \citep{sohn2023styledrop}, DEADiff \citep{qi2024deadiff}, CSGO \citep{xing2024csgo} and InstantStyle \citep{wang2024instantstyle, wang2024instantstyleplus}. The details of these comparisons, along with the experimental setup and metrics, are described in the \aref{sec:exp_details}.

\subsection{Effectiveness of CFG and \NVQG}
\label{exp:cfgnqg}
This section analyzes the effects of Classifier-Free Guidance (CFG) and \NegativeQueryGuidance~(\NVQG) on image generation, with a focus on text alignment and content leakage. \fref{fig:cfg_nqg} shows the results of the three configurations.
\textbf{In the 1\textsuperscript{st} row}, without CFG and \NVQG, the generated images suffer from severe artifacts. The absence of CFG causes poor image quality resulting in significant misalignment with the prompt. 
\textbf{In the 2\textsuperscript{nd} row}, CFG with \swappingselfattention{} improves the text misalignment by boosting image quality. Here, the ``cat'' in target text prompt becomes clearer in the generated images. However, content leakage from the reference image remains where unwanted elements (layouts, structure, and composition) from the reference image affect the results. 

\textbf{In the 3\textsuperscript{rd} row}, \NVQG{} releases the content leakage and produces the best results closely matching the text prompt while reflecting style reference. 

Overall, \fref{fig:cfg_nqg} demonstrates the critical role of \NVQG{} in reducing content leakage, enjoying the quality boosting of CFG. Together, they ensure that the generated images align to both the target text prompt and the visual style prompts, resulting in high-quality, coherent outputs.

We qualitatively showcase diversity of results within a text prompt in \fref{afig:more_diversity_results} and text alignment with complex text prompts in \fref{fig:complex_prompt}.

\subsection{Analysis for Swapping Self-Attention}
\label{sec:exp_swap_sa}
\paragraph{Optimal layers in upblocks}Since recent large T2I DMs consist of many layers, we further analyze the behavior by changing the start of the swapping while the end of the swapping is fixed at the end.
We use four key metrics as
shown in \fref{fig:ablation:upblock}, there is a layer where all four metrics abruptly change (red line). Notably, this point remains consistent regardless of the reference image. All the results in the paper use only one fixed optimal layer.

We choose this layer as the optimal start of the swapping for a balanced trade-off among all aspects. We provide qualitative results with detailed split of layers in \fref{afig:layer_ablation_qual_anime} and \dref{afig:layer_ablation_qual_lie_art}. 

\begin{figure}[t]
    \centering
    \includegraphics[width=0.8\linewidth]{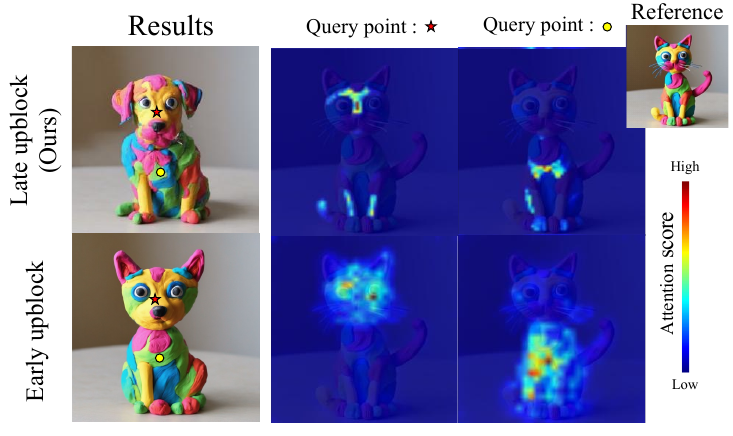}
    \caption{\textbf{Attention map visualization over late and early upblock layers.}
    The late upblock better focuses on the style-corresponding region than the early upblock, leading to more freedom to reassemble small parts. The early upblock attends a larger region leading to content leakage.
    }
    \label{fig:ablation:attentionmap}
\end{figure}

\paragraph{Visualizing Attention maps}\fref{fig:ablation:attentionmap} compares average attention maps from the late upblock and the early upblock applying swapping self-attention.
Using late upblock has more freedom to reassemble the reference style elements leading to more doggy results than early upblock which produces some cat-like attributes.
The right two columns visualize the attention weight of query points marked as red stars and yellow dots.
Swapping self-attention on late upblock reassembles features from a style correspondence, e.g., texture and color.
On the other hand, swapping self-attention on early upblock reassembles features from a wider area with different styles. This comparison clarifies the reasons for using only late upblock. Please refer to \fref{afig:ablation:attentionmap}, \ref{afig:layer_ablation_visual_advanced} for a more visualization and a detailed analysis.
\begin{figure}[t]
    \centering
    \includegraphics[width=0.9\linewidth]{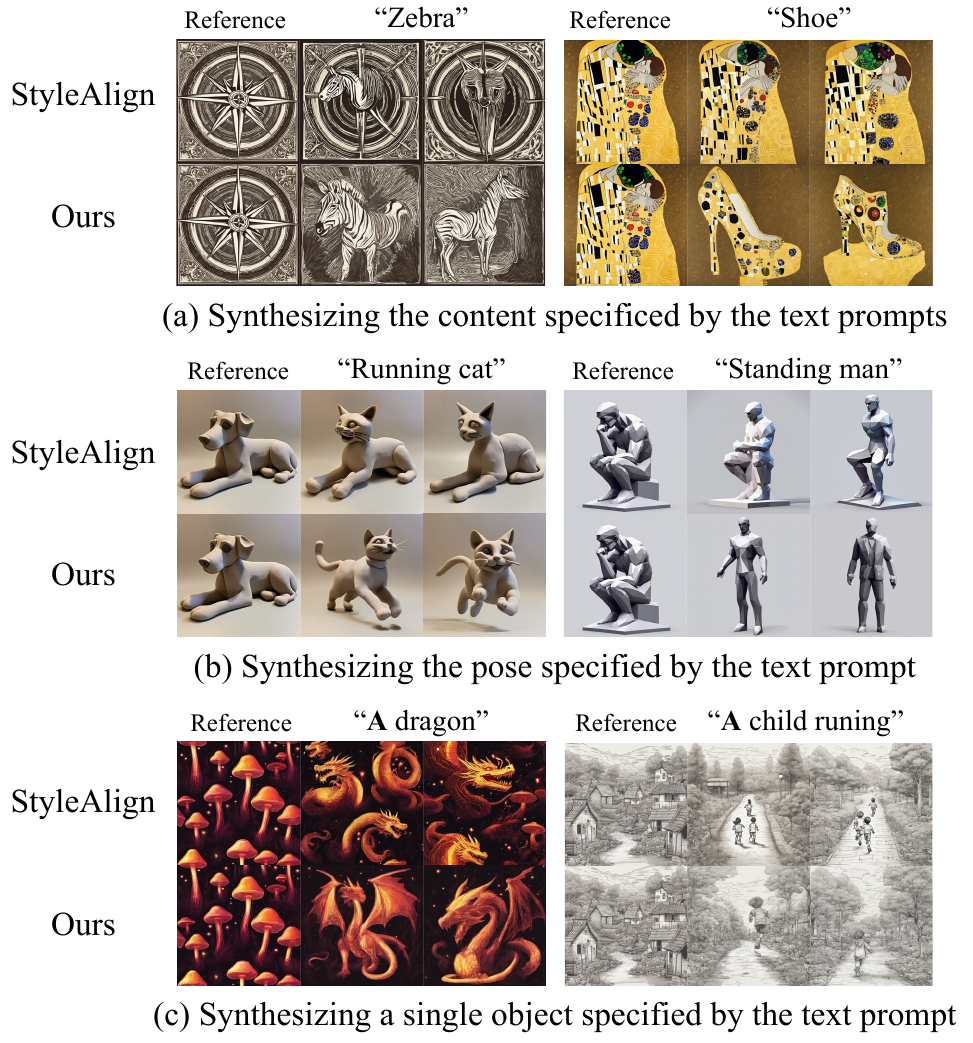}
    
    \caption{
    \textbf{Comparison for content leakage.} While StyleAligned suffers from content leakage from the reference, our results clearly align with the text prompts
    }
    \label{fig:ablation:vs_SA_senario}
\end{figure}

\begin{figure*}[t]
    \centering
    \includegraphics[width=1.0\textwidth]{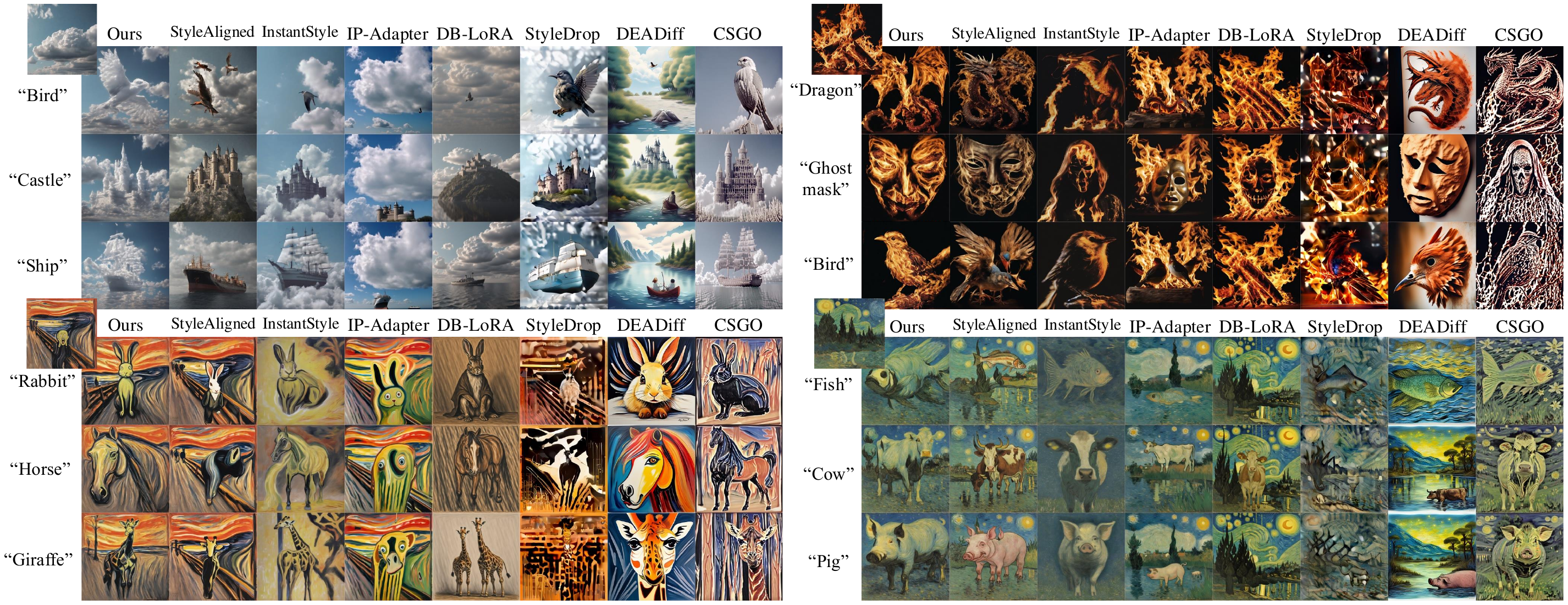}
    \caption{\textbf{Qualitative comparison across various styles and text prompts.} \Ours{} faithfully reflects style elements in reference images without causing content leakage from the reference images.}
    \label{fig:vs_competitor}
\end{figure*}

\begin{figure}[t]
    \centering
        \centering
        \vspace{-2mm}
        \includegraphics[width=1.0\linewidth]{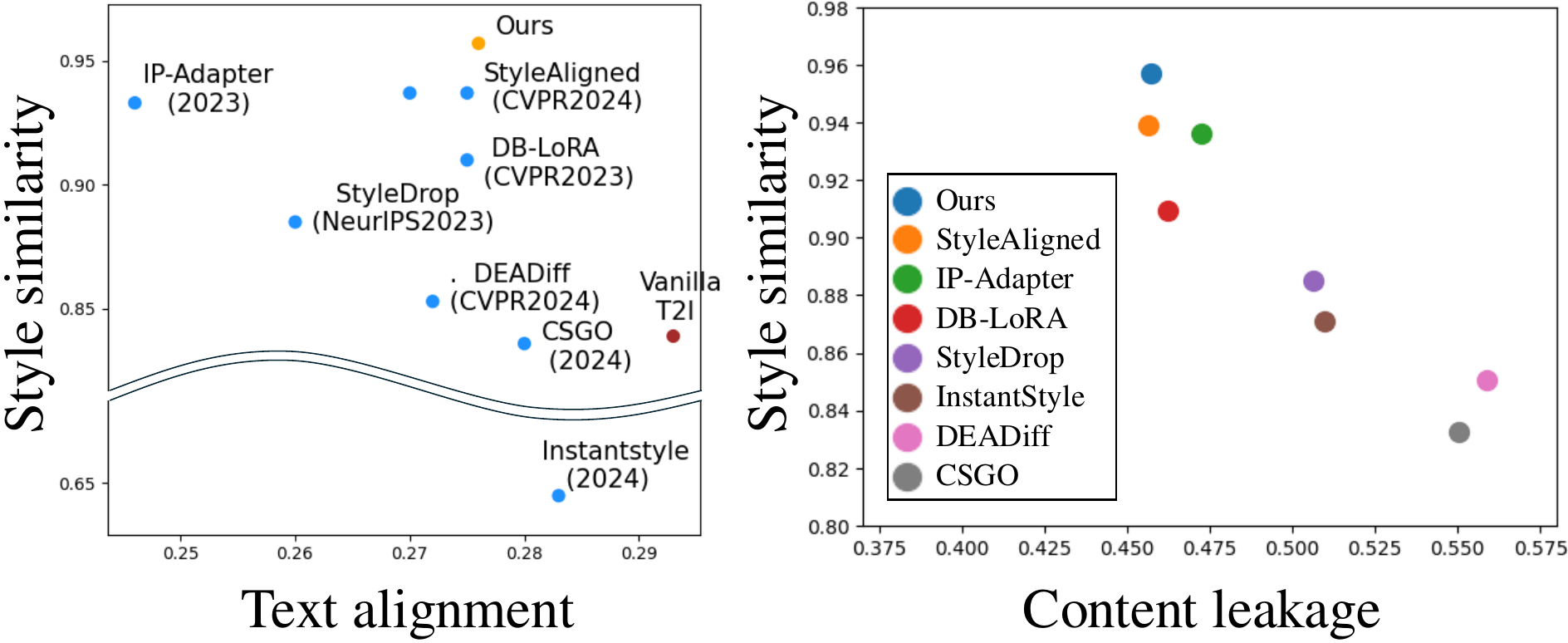}
        \caption{\textbf{Quantitative comparison}.
        Ours achieves the best style similarity while not compromising text alignment and being free from content leakage.}
        \vspace{-5mm}
        
        \label{fig:vs_competitors_quan}
\end{figure}

\subsection{Comparison against competitors}
\label{sec:exp_comparison_competitors}

\paragraph{Style \& content control}

We provide a qualitative comparison in \fref{fig:vs_competitor}, focusing on controlling style and content. Our method faithfully synthesizes content from the text prompt with the style of the reference image. In contrast, other methods add elements like color or texture not in the reference (e.g., feathers, bricks, iron, skin) and often suffer from content leakage (e.g., layout, screaming person, castle), which compromises text prompt faithfulness. Quantitative results in \fref{fig:vs_competitors_quan} support these findings: IP-Adapter shows higher style similarity but neglects text prompts significantly.

\paragraph{Diversity within a text specification}
Starting from different initial noises, the diffusion models trained on a large dataset produce diverse results within the specification of a text prompt. \fref{fig:diversity_vs_competitors}, \ref{afig:content_leakage_deadiff} shows that ours can synthesize various poses and viewpoints while others barely change, i.e., other methods limit the diversity of the pre-trained model due to content leakage. We further support our diversity in \fref{afig:more_diversity_results}.

\paragraph{Content leakage} Content leakage refers to the phenomenon where the content of a reference image appears in a result. As evident in \fref{fig:vs_competitor_frog}, our model exhibits significantly less content leakage when compared to other models. We focus on a comparison against a runner-up method, StyleAligned, particularly in terms of how content leakage can be an obstacle to controlling the content using a text prompt. \fref{fig:ablation:vs_SA_senario}a compares examples where strong content leakage prevents the object from the text from appearing in StyleAligned. We often observe famous paintings in the reference easily leak into the results of StyleAligned while ours does not struggle. \fref{fig:ablation:vs_SA_senario}b compares examples where the pose in the reference leaks into the results of StyleAligned preventing the reflection of specified pose in the text prompt. On the other hand, our method reflects the correct poses. \fref{fig:ablation:vs_SA_senario}c compares examples where the number of small instances in the reference leaks into the result of StyleAligned. Contrarily, our method correctly synthesizes a single instance of the content specified by the text. In \fref{fig:vs_competitors_quan}, we provide quantitative comparison results. It supports that \Ours{} achieves the best style similarity while not suffering content leakage. We can also observe that models with high content leakage exhibit poor text alignment.

\paragraph{Preserving the content of the original denoising process}
\fref{afig:ablation:init_noise_is_content} shows the results of ours and other methods using the same initial noise in each column. Our method precisely reflects the style in the reference with minimal changes of contents in the original denoising process. On the other hand, the other methods severely alter pose, shape, or layouts due to content leakage. It is an important virtue of controlling style to keep the rest intact. This also supports that ours is free from content leakage during the denoising process. We provide more results in \fref{afig:ablation:vs_SA_senario}. 

\begin{figure*}[t]
    \centering
    \includegraphics[width=1.0\linewidth]{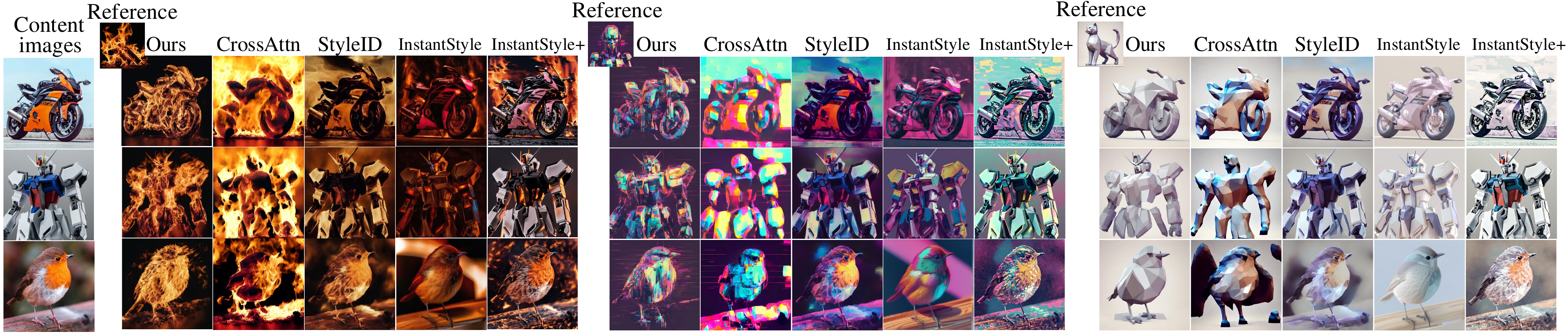}
    \vspace{-1.5mm}
    \caption{\textbf{Qualitative comparison in I2I style transfer task.} We compare our method for I2I style transfer task where the content image is given. Compared to the previous methods, ours achieves the best style similarity.} 
    \label{fig:controlnet_comp}
    \vspace{-2mm}
\end{figure*}

\begin{figure}[t]
    \centering
    \vspace{-2mm}
    \includegraphics[width=0.85\linewidth]{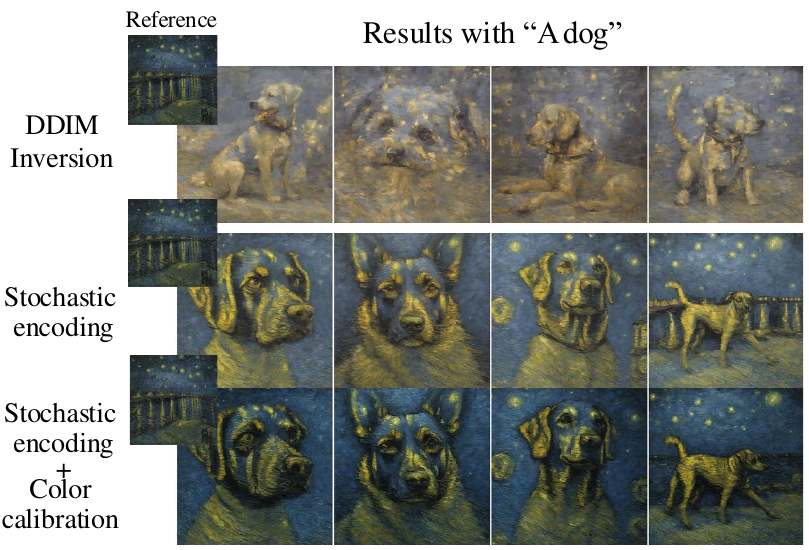}
    \caption{\textbf{Comparison of DDIM inversion vs. stochastic encoding and the effect of color calibration.} Stochastic encoding reduces artifacts in the resulting images, while color calibration better reflects the colors of the reference image. } 
    \label{fig:vsp_add_noise_calibration_starry-night}
    \vspace{-2mm}
\end{figure}


\subsection{DDIM inversion vs. Our strategy}
\label{sec:exp_real_image}
\begin{figure}[t]
    \centering
    \includegraphics[width=1.0\linewidth]{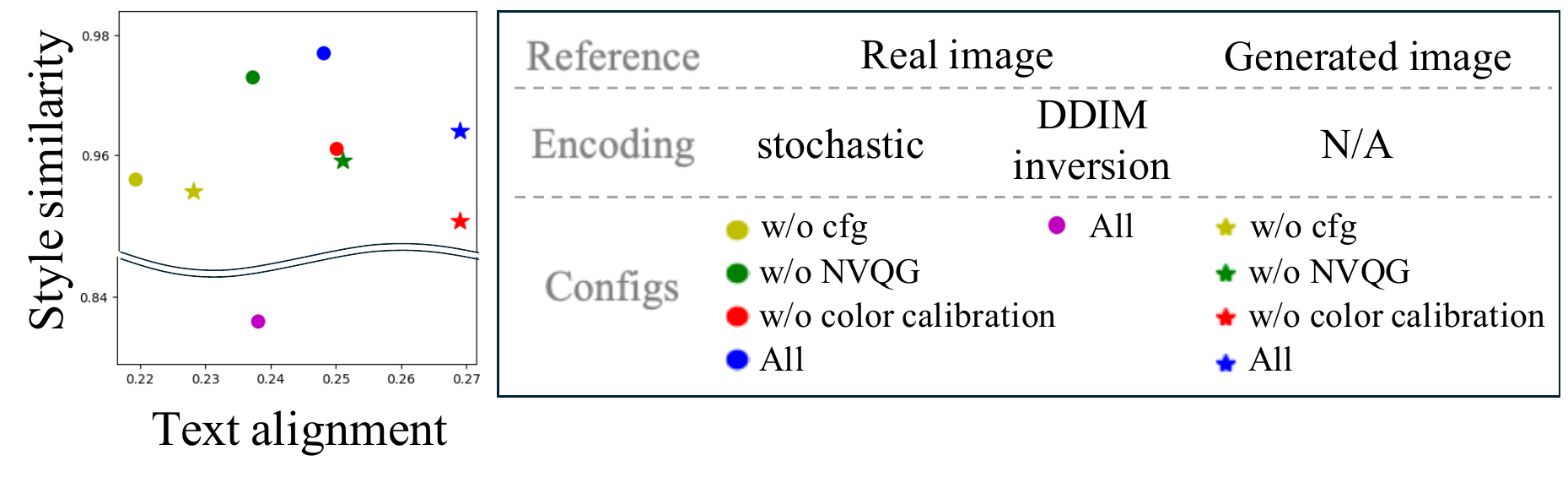}
    \caption{\textbf{Ablation study of four proposed methods with a real reference and a generated reference setting.}. We evaluated performance by individually removing each of the four proposed methods. In all cases, both generated and real images showed similar improvements in style similarity and text alignment. Moreover, stochastic encoding outperformed DDIM inversion in all aspects, comparing purple and blue points.}
    \label{afig:stochastic_calib_real_generate}
\end{figure}

\fref{fig:vsp_add_noise_calibration_starry-night} shows that \ours{} can take real images as style reference with our strategies. As shown in \fref{fig:vsp_add_noise_calibration_starry-night}, stochastic encoding outperforms DDIM inversion and color calibration improves text alignment and style similarity. We provide more results in \fref{afig:vsp_add_noise_calibration_six_exp}. Moreover, we show that our strategy can boost the performance of the other self-attention variants \cite{hertz2023style, cao_2023_masactrl} in \fref{afig:ablation:add_noise_with_SA}, \ref{afig:ablation:add_noise_with_masactrl} and color calibration can be used for generation settings in \fref{afig:color_calibration_vsp}.
We provide an ablation study in \fref{afig:stochastic_calib_real_generate} for each configuration (swapping self-attention, NVQG, and color calibration) in both real reference and generated reference settings. In both settings, swapping self-attention and NVQG improve style similarity and text alignment, while color calibration helps improve style similarity. Additionally, stochastic encoding outperforms DDIM inversion.


\subsection{Comparison for I2I style transfer}
\label{sec:comp_i2i_scenario}
In \fref{fig:controlnet_comp}, we present a qualitative comparison between our method using ControlNet and existing state-of-the-art methods, CrossAttn~\citep{alaluf2024cross} and StyleID~\citep{chung2024style}, for the I2I style transfer task where a content image is provided. Both CrossAttn and StyleID inject the key \& value obtained by inversion from the style images.

CrossAttn often fails to transfer the detailed representation of the reference style, leading to a rough, blocky appearance. For instance, the center examples in \fref{fig:controlnet_comp} show that the style details are absent or inaccurate compared to our results. Similarly, StyleID suffers poor style reflection and color deficiency in the center and the rightmost example. In contrast, ours effectively reflects the details of the reference style image, with no noticeable transfer of the color values from the content images.

\section{Conclusion and Limitation}

In this paper, we introduce \ours{} which faithfully applies the style of real reference images without content leakage in a training-free manner. \textbf{CFG with swapping self-attention} captures the reference image’s style accurately and allows for direct content generation from the text. This integration with CFG enhances performance by balancing content generation and style transfer. To address content leakage, we propose \textbf{Negative visual query guidance} that ensures the reference image’s content does not interfere with the text-specified content, enhancing diversity and improving text alignment \& style similarity. Additionally, \textbf{Stochastic encoding} efficiently maps real images to suitable latents, removing artifacts. \textbf{Color calibration} aligns the final output to the reference’s color statistics.

Our method shows both qualitative and quantitative improvements over existing approaches, providing a robust solution for visual style prompting without complex training. We also analyze where to apply swapping self-attention, identifying optimal layers to balance style transfer and content fidelity. \ours{} outperforms prior methods and works well with algorithms like ControlNet~\citep{zhang2023adding} and DB-LoRA~\citep{lora_repo}, as shown in \fref{fig:controlnet_db_lora}.

However, \ours{} is limited by the capabilities of the pretrained diffusion models and cannot generate elements outside their scope (e.g., ``stone golem'' in \fref{fig:limit}a). Also, if visual and textual styles conflict, the visual style dominates (\fref{fig:limit}b). Future work could explore extending our method to other domains such as video, expanding its applicability and research potential.



\paragraph{Acknowledgements}
We thank Jin-Hwa Kim for helpful comments. All experiments were conducted on NAVER Smart Machine Learning (NSML) platform~\cite{kim2018nsml, sung2017nsml}. This work was supported by Institute of Information \& communications Technology Planning \& Evaluation (IITP) grant funded by the Korea government (MSIT) (No. RS-2024-00439762, Developing Techniques for Analyzing and Assessing Vulnerabilities, and Tools for Confidentiality Evaluation in Generative AI Models).

{
    \small
    \bibliographystyle{ieeenat_fullname}
    \bibliography{main}
}

\clearpage
\setcounter{page}{1}
\setcounter{section}{0}
\maketitlesupplementary

\section{Related work}
\label{sec:relatedwork}
\subsection{Controlling style with training}
Dreambooth~\citep{ruiz2023dreambooth, kumari2023multi, everaert2023diffusion, lora_repo} and Adapter~\citep{ zhang2023adding, li2023gligen, sohn2023styledrop} variants fine-tune a pre-trained diffusion model (DM) with a few images with a same style to completely change the model. Textual inversion variants~\citep{gal2022image, han2023highly} learn a customized text embedding to use the same model with a minimal extra component. 
Meanwhile, ControlNet~\citep{zhang2023adding} stands out as one of the most effective frameworks to guide structural contents. Similarly, adapter-based methods \citep{ye2023ip, wang2023styleadapter} attach an auxiliary image encoder which receives style reference. Several works\citep{vsubrtova2023analogies,sun2023imagebrush,nguyen2023visual, xing2024csgo, qi2024deadiff} train the model to solve image analogy where instructions demonstrate the desired manipulation. 

Although these methods often succeed in controlling style, they require a time-consuming additional training procedure or limit the applicable styles within their training set, which hinders practical usage.


\subsection{Training-free control with feature manipulation}
Manipulating the intermediate features in DMs inherently changes the resulting images even with a frozen DM. 

\citet{alaluf2024cross} and \citet{chung2024style} transfer the style of an image to another image via self-attention with DDIM inversion achieving insufficient control. Furthermore, their main goal is image-to-image (I2I) rather than text-to-image (T2I); naive two-stage T2I2I struggles in trade-off between style reflection and content preservation. StyleAligned \citep{hertz2023style} similarly shares self-attention between two processes but fails to exclude the original style elements because it keeps the original features. 
Meanwhile, sharing self-attention features improves temporal appearance consistency over multiple frames~\citep{wu2023tune, yang2023rerender} in video editing task.

Moving forward, we unveil the multiple missing components for accurate control over style and content, producing outstanding results. 

On the other hand, Plug-n-Play~\citep{tumanyan2023plug} and MasaCtrl~\citep{cao_2023_masactrl} inject self-attention features from one process to another to convey \textit{structure} and \textit{object}, respectively, of the first process, rather than conveying style elements.

\subsection{Inverting images to noise for real image editing in diffusion models}
Training-free editing methods in the previous subsection manipulate the features of attention layers. If synthetic images with known latent noises are used as reference images, this process is straightforward. However, when editing with real images, the methods require inverting the images into the initial latent noise maps. A majority of diffusion-based editing work \citep{alaluf2024cross, hertz2022prompt, mokady2023null, couairon2022diffedit, wallace2023edict} and self-attention variants \citep{hertz2023style, cao_2023_masactrl, alaluf2024cross} employs DDIM inversion \citep{song2020denoising} for its deterministic mapping between noise and image. However, DDIM inversion often suffers error at each step from $z_{t-1}$ to $z_t$ and inverted noise from a real image through DDIM does not follow the standard Gaussian distribution \citep{mokady2023null, miyake2025negative, han2023improving, garibi2024renoise}. 

DDPM sampling is an alternative that produces the standard Gaussian noise \citep{ho2020denoising,huberman2024edit, cyclediffusion, meng2022sdedit}. However, they require lengthy iterative sampling for inversion or do not distinguish the reference denoising process from the inference denoising process.
Compared to these inverting methods, our stochastic encoding consists of a one-step operation and produces statistically aligned intermediate latents. 




\renewcommand{\thetable}{A\arabic{table}}
\renewcommand{\thefigure}{A\arabic{figure}}
\setcounter{figure}{0}
\setcounter{table}{0}

\section{Experiments details and metrics}
\label{sec:exp_details}
\paragraph{Details} 
We use SDXL \cite{podell2023sdxl} as our pretrained text-to-image diffusion model and choose the $\text{24}^\text{th}$ layer and the after. We also validate our methods on Stable diffusion (SD) v1.5 \cite{rombach2022high}. Results with SD v1.5 are shown in \fref{afig:SDv15}. We set classifier-free guidance as 7.0 and run DDIM sampling with 50 timesteps following the typical setting. 
The initial noises for a text prompt are identical across competitors for fair comparison.
We use the official implementation of IP-Adapter and StyleAligned. For Dreambooth-LoRA\footnote{https://huggingface.co/docs/diffusers/training/lora}
, we use diffusers pipeline provided by Huggingface. Since the official code of StyleDrop is not available, we use an unofficial implementation\footnote{https://github.com/huggingface/diffusers/tree/main/examples/amused} provided by Huggingface. All competitors are based on SDXL except StyleDrop (unofficial MUSE).
As Dreambooth-LoRA, a training-based approach, requires multiple images, we train the models with five images: the original image and quarter-split patches of the reference image, because using only one image usually leads to destructive results or suffers from overfitting.
For IP-adapter, we choose $\lambda{}=0.5$ which is the best weighting factor for the task. For the visualization of the attention maps in \fref{fig:ablation:attentionmap}, we average the multi-head attention maps all together along the channel axis at the 20th denoising timestep.

\paragraph{Metrics}
Following \cite{voynov2023p+, ruiz2023dreambooth}, we use DINO (ViT-B/8) embeddings \cite{caron2021emerging} to measure style similarity between a reference image and a resulting image. In addition, we provide a quantitative comparison with Gram matrix \cite{gatys2015neural} to assess style similarity in \tref{tab:gram}. We use CLIP (ViT-L/14) embeddings \cite{radford2021learning} to measure the alignment between text prompts and resulting images. We use LPIPS \cite{zhang2018unreasonable} to measure diversity by average LPIPS between different resulting images in the same text prompt.
We use Kolmogorov-Smirnov test \cite{16e7f618-c06b-3d10-8705-1086b218d827} to measure gaussianity. For quantitative evaluation and comparison, we prepare 720 synthesized images from 40 reference images, 120 content text prompts (3 contents per 1 reference), and 6 initial noises. The reference images are generated from 40 stylish text prompts. \aref{asec:style-content-prompt-list} provides the text prompt set. We also conducted a user study to evaluate the effectiveness of our method in \aref{sec:user_study}.

\begin{figure}[t]
    \centering
    \includegraphics[width=1.0\linewidth]{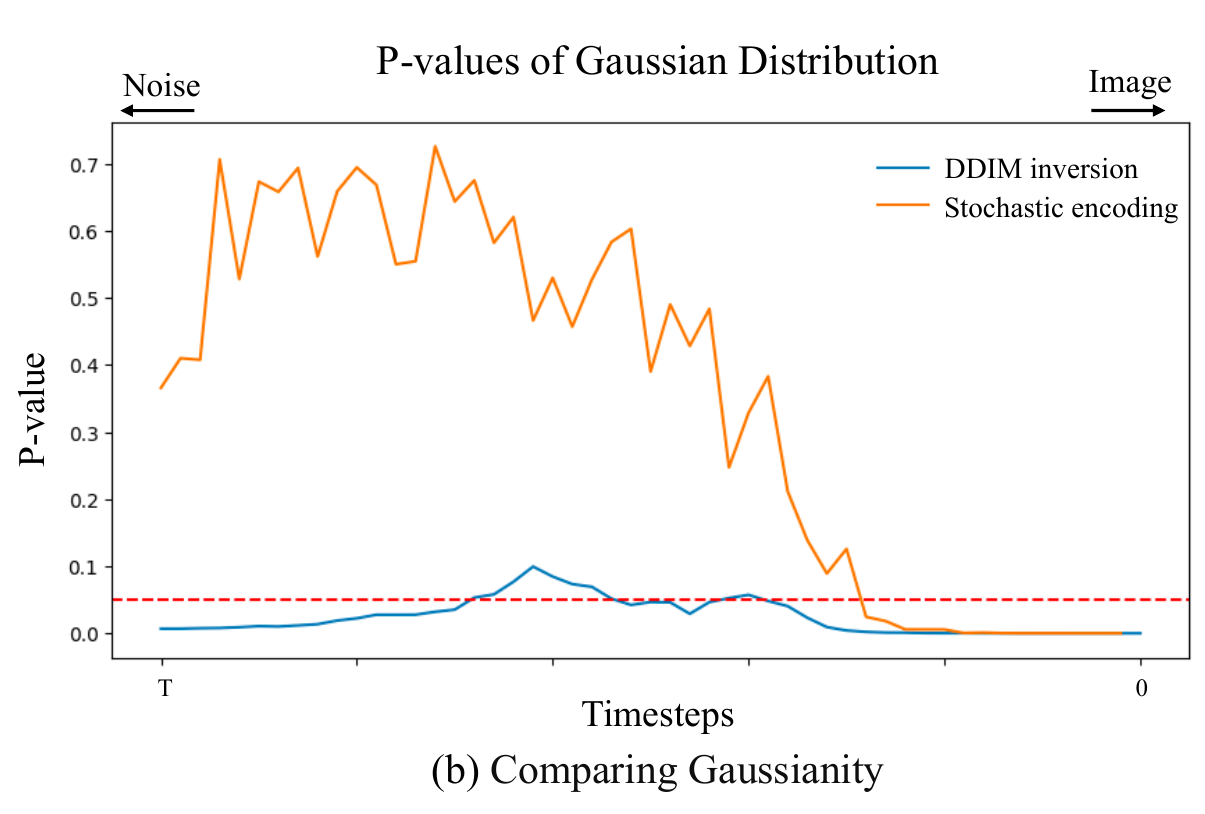}
    \caption{\textbf{Stochastic encoding produces the latents closer to the standard Gaussian distribution compared to DDIM inversion.} A P-value above 0.05 suggests that the data likely follows the standard Gaussian distribution. 
} 
    \label{afig:p-values_of_gaussian}
\end{figure}

\begin{table}[h]
    \centering
    \caption{
    \textbf{Quantitative comparison.} We compare the results for style similarity by utilizing a gram matrix.
    }
    \begin{tabular}{ccccc}
    \toprule
    Ours & StyleAligned & IP-Adapter & DB-LoRA & StyleDrop \\
    \midrule
    \textbf{0.791} & 0.759 & 0.768 & 0.759 & 0.659 \\
    \bottomrule
    \end{tabular}
    \label{tab:gram}
\end{table}

\onecolumn

\newpage

\section{Algorithm: Visual style prompting with a real image as a reference }
\label{sec:algorithm}
\begin{algorithm}[H]
\SetAlgoLined
\KwIn{Reference latent $x_0^{visual}$, number of diffusion steps $T$, color calibration range $[t_{\text{start}}, t_{\text{end}}]$, precomputed constants $\alpha_t$, noise scale $\sigma_t$, model $\epsilon_\theta$}
\KwOut{Denoised latent $x_0$}

\SetKwFunction{FMain}{color\_calibration}
\SetKwProg{Fn}{Function}{:}{}
\Fn{\FMain{$x_t$, $\hat{x}_t$, $x_0^{visual}$}}{
    $x_{pred} \gets \frac{x_t - \sqrt{1 - \alpha_t} \cdot \epsilon_\theta(\hat{x}_t)}{\sqrt{\alpha_t}}$ \tcp*{predicted $x_0$}
    $x_{dir} \gets \sqrt{1 - \alpha_{t-1} - \sigma_t^2} \cdot \epsilon_\theta(\hat{x}_t)$ \tcp*{direction pointing to $x_t$}
    $\epsilon \sim \mathcal{N}(0, I)$ \tcp*{Generate random noise}
    $x_{noise} \gets \sigma_t \cdot \epsilon$ \tcp*{random noise}
    $\hat{x}_{pred} \gets \text{adain}(x_{pred}, x_0^{visual})$ \tcp*{calibrated predicted $x_0$}
    $x_{t-1} \gets \sqrt{\alpha_{t-1}} \cdot \hat{x}_{pred} + x_{dir} + x_{noise}$ \tcp*{updated $x_{t-1}$}
    \KwRet $x_{t-1}$
}

\SetKwFunction{FAdaIN}{adain}
\SetKwProg{Fn}{Function}{:}{}
\Fn{\FAdaIN{$x, x^{visual}$}}{
    $\mu_x, \sigma_x \gets \text{channel\_wise\_mean\_std}(x)$ \tcp*{mean and std of $x$}
    $\mu_x^{visual}, \sigma_x^{visual} \gets \text{channel\_wise\_mean\_std}(x^{visual})$ \tcp*{mean and std of $x^{visual}$}
    $x_{norm} \gets \frac{x - \mu_x}{\sigma_x}$ \tcp*{normalize $x$}
    $x_{adain} \gets \sigma_x^{visual} \cdot x_{norm} + \mu_x^{visual}$ \tcp*{scale and shift}
    \KwRet $x_{adain}$
}

\SetKwFunction{FForward}{stochastic\_encoding}
\SetKwProg{Fn}{Function}{:}{}
\Fn{\FForward{$x$, $t$}}{
    $\epsilon \sim \mathcal{N}(0, I)$ \tcp*{Generate random noise}
    $x_t \gets \sqrt{\alpha_t} \cdot x + \sqrt{1 - \alpha_t} \cdot \epsilon$ \;
    \KwRet $x_t$
}

Initialize $x_T \gets \mathcal{N}(0, I)$\;
Initialize $x_T^{visual} \gets \text{stochastic\_encoding}(x_0^{visual}, T)$\;
Initialize $t \gets T$\;

\For{$t = T$ \KwTo $1$}{
    $\hat{x}_t \gets \text{CFG\_NQG\_with\_swapping\_self\_attention}(x_t, x_t^{visual})$\tcp*{(Fig 2)In denoising process, $x_t$ and $x_t^{visual}$ are swapped}
    \tcp*{and predict $\hat{\epsilon}_t$ }
    \If{$t_{\text{start}} \leq t \leq t_{\text{end}}$}{
        $x_{t-1} \gets \text{color\_calibration}(x_t,\hat{x}_t, x_0^{visual})$ \;
    }
    \Else{
        $x_{pred} \gets \frac{\hat{x}_t - \sqrt{1 - \alpha_t} \cdot \epsilon_\theta(\hat{x}_t)}{\sqrt{\alpha_t}}$ \tcp*{predicted $x_0$}
        $x_{dir} \gets \sqrt{1 - \alpha_{t-1} - \sigma_t^2} \cdot \epsilon_\theta(\hat{x}_t)$ \tcp*{direction pointing to $x_t$}
        $\epsilon_t \sim \mathcal{N}(0, I)$ \tcp*{Generate random noise}
        $x_{noise} \gets \sigma_t \cdot \epsilon_t$ \tcp*{random noise}
        $x_{t-1} \gets \sqrt{\alpha_{t-1}} \cdot x_{pred} + x_{dir} + x_{noise}$ \tcp*{updated $x_{t-1}$}
    }
    Decrease $t$ by 1\;
}

\KwRet $x_0$
\caption{Visual style prompting with a real image as a reference}
\label{alg:stochastic_encoding_color_calibration}
\end{algorithm}

\newpage

\section{Style-Content prompt list}
\label{asec:style-content-prompt-list}
1. the great wave off kanagawa in style of Hokusai: (1) book (2) cup (3) tree 
 
2. fire photography, realistic, black background: (1) a dragon (2) a ghost mask (3) a bird 
 
3. A house in stickers style.: (1) A temple (2) A dog (3) A lion 
 
4. The persistence of memory in style of Salvador Dali: (1) table (2) ball (3) flower 
 
5. pop Art style of A compass . bright colors, bold outlines, popular culture themes, ironic or kitsch: (1) A violin (2) A palm tree (3) A koala 
 
6. A compass rose in woodcut prints style.: (1) A cactus (2) A zebra (3) A blizzard 
 
7. A laptop in post-modern art style.: (1) A man playing soccer (2) A woman playing tennis (3) A rolling chair 
 
8. A horse in colorful chinese ink paintings style: (1) A dinosaur (2) A panda (3) A tiger 
 
9. A piano in abstract impressionism style.: (1) A villa (2) A snowboard (3) A rubber duck 
 
10. A teapot in mosaic art style.: (1) A kangaroo (2) A skyscraper (3) A lighthouse 
 
11. A robot in digital glitch arts style.: (1) A cupcake (2) A woman playing basketball (3) A sunflower 
 
12. A football helmet in street art graffiti style.: (1) A playmobil (2) A truck (3) A watch 
 
13. Teapot in cartoon line drawings style.: (1) Dragon toy (2) Skateboard (3) Storm cloud 
 
14. A flower in melting golden 3D renderings style and black background: (1) A piano (2) A butterfly (3) A guitar 
 
15. Slices of watermelon and clouds in the background in 3D renderings style.: (1) A fox (2) A bowl with cornflakes (3) A model of a truck 
 
16. pointillism style of A cat . composed entirely of small, distinct dots of color, vibrant, highly detailed: (1) A lighthouse (2) A hot air balloon (3) A cityscape 
 
17. the garden of earthly delights in style of Hieronymus Bosch: (1) key (2) ball (3) chair 
 
18. Photography of a Cloud in the sky, realistic: (1) a bird (2) a castle (3) a ship 
 
19. A mushroom in glowing style.: (1) An Elf (2) A dragon (3) A dwarf 
 
20. The scream in Edvard Munch style: (1) A rabbit (2) a horse (3) a giraffe 
 
21. the girl with a pearl earring in style of Johannes Vermeer: (1) door (2) pen (3) boat 
 
22. A wild flower in bokeh photography style.: (1) A ladybug (2) An igloo in antarctica (3) A person running 
 
23. low-poly style of A car . low-poly game art, polygon mesh, jagged, blocky, wireframe edges, centered composition: (1) A tank (2) A sofa (3) A ship 
 
24. Kite surfing in fluid arts style.: (1) A pizza (2) A child doing homework (3) A person doing yoga 
 
25. anime artwork of cat . anime style, key visual, vibrant, studio anime, highly detailed: (1) A lion (2) A chimpanzee (3) A penguin 
 
26. A cactus in mixed media arts style.: (1) A shopping cart (2) A child playing with cubes (3) A camera 
 
27. the kiss in style of Gustav Klimt: (1) shoe (2) cup (3) hat 
 
28. Horseshoe in vector illustrations style.: (1) Vintage typewriter (2) Snail (3) Tornado 
 
29. play-doh style of A dog . sculpture, clay art, centered composition, Claymation: (1) a deer (2) a cat (3) an wolf 
 
30. A cute puppet in neo-futurism style.: (1) A glass of beer (2) A violin (3) A child playing with a kite 
 
31. the birth of venus in style of Sandro Botticelli: (1) lamp (2) spoon (3) flower 
 
32. line art drawing of an owl . professional, sleek, modern, minimalist, graphic, line art, vector graphics: (1) a Cheetah (2) a moose (3) a whale 
 
33. origami style of Microscope . paper art, pleated paper, folded, origami art, pleats, cut and fold, centered composition: (1) Giraffe (2) Laptop (3) Rainbow 
 
34. A crystal vase in vintage still life photography style.: (1) A pocket watch (2) A compass (3) A leather-bound journal 
 
35. The Starry Night, Van Gogh: (1) A fish (2) A cow (3) A pig 
 
36. A village in line drawings style.: (1) A building (2) A child running in the park (3) A racing car 
 
37. A diver in celestial artworks style.: (1) Bowl of fruits (2) An astronaut (3) A carousel 
 
38. A frisbee in abstract cubism style.: (1) A monkey (2) A snake (3) Skates 
 
39. Flowers in watercolor paintings style.: (1) Golden Gate bridge (2) A chair (3) Trees (4) An airplane 
 
40. A horse in medieval fantasy illustrations style.: (1) A castle (2) A cow (3) An old phone 

\newpage

\section{User study}
\label{sec:user_study}

\begin{figure}[h]
    \centering
    \includegraphics[width=1.0\textwidth]{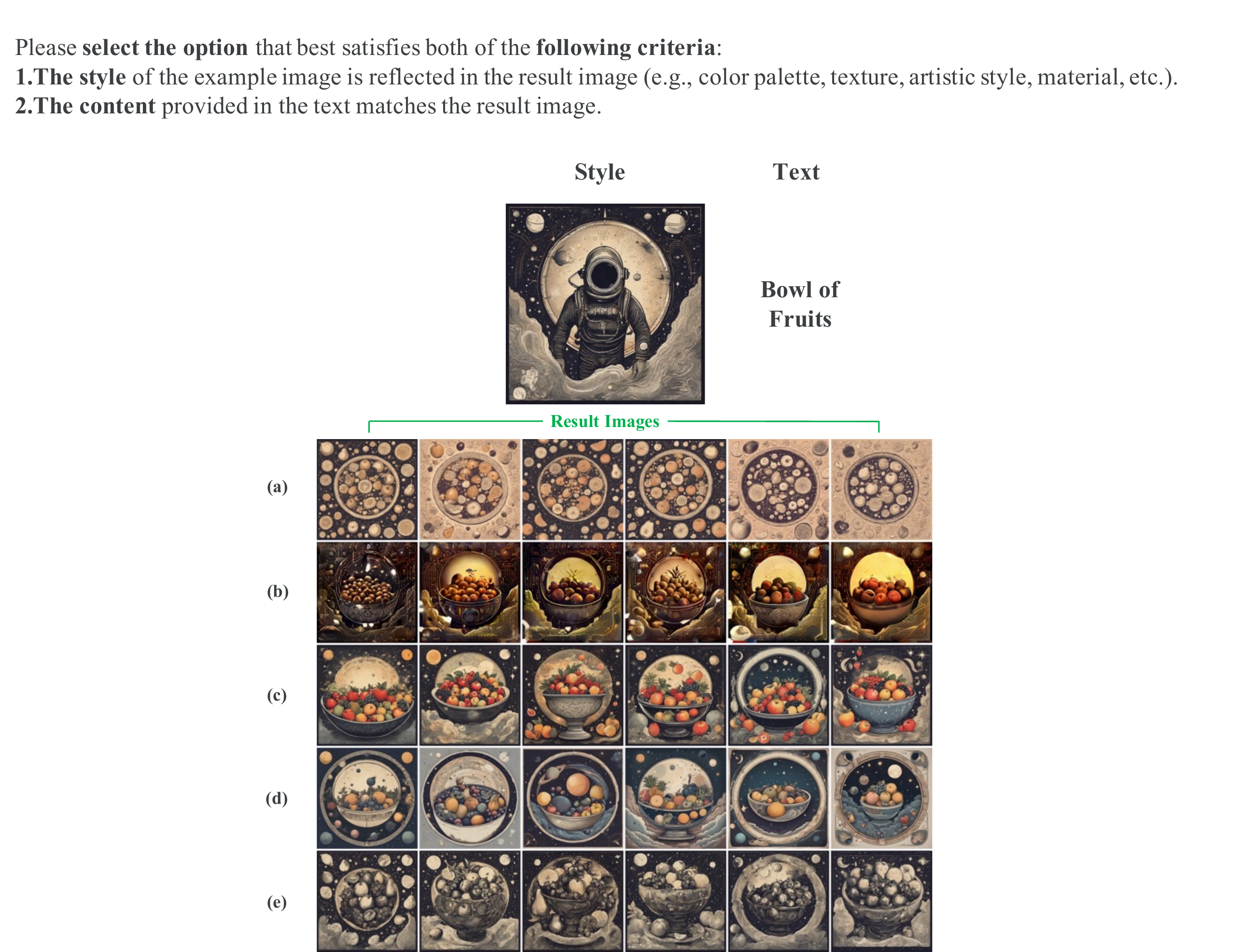}
    \caption{\textbf{Example of a user study.} Each row of images represents the result obtained by different method. The user had to assess
which row is better in terms of style alignment and text alignment.}
    \label{afig:user_study_ex}
\end{figure}

\begin{figure}[h]
    \centering
    \includegraphics[width=1.0\textwidth]{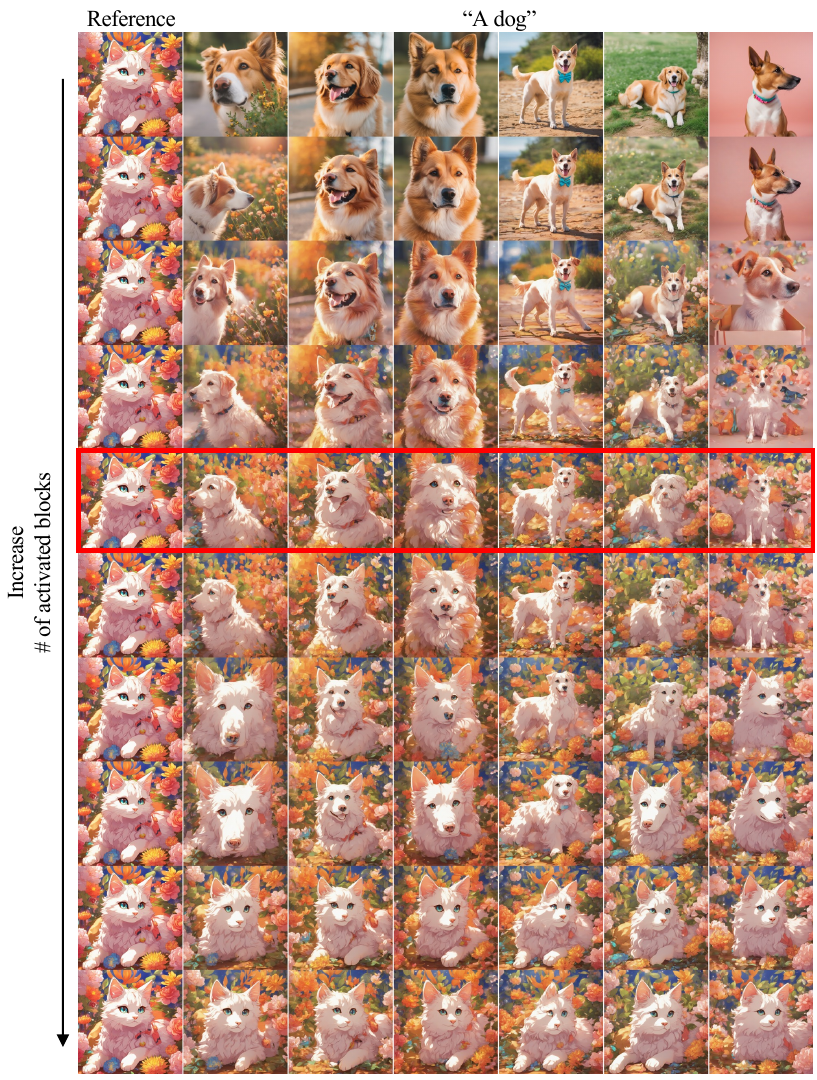}
    \caption{Selective \crossstyleattention{} is important to avoid content leakage while preserving style similarity. Content leakage decreases diversity and text alignment.}
    \label{afig:layer_ablation_qual_anime}
\end{figure}

\begin{figure}[h]
    \centering
    \includegraphics[width=1.0\textwidth]{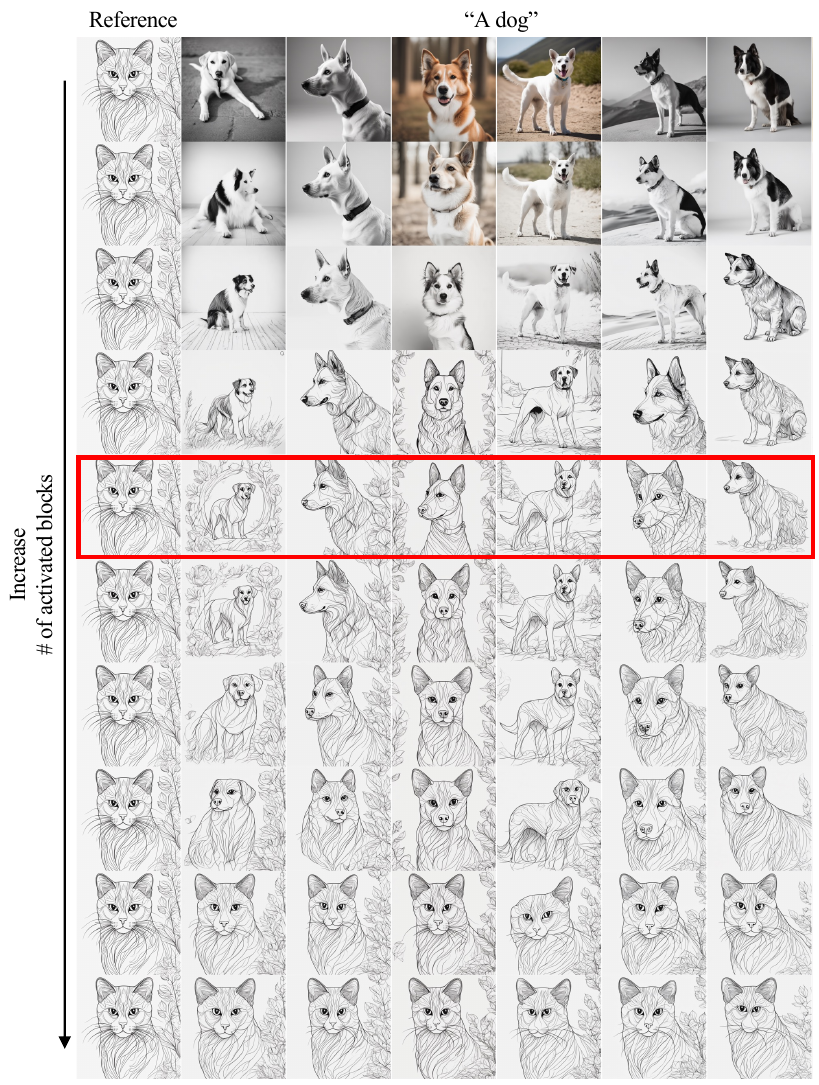}
    \caption{Selective \crossstyleattention{} is important to avoid content leakage while preserving style similarity. Content leakage decrease diversity and text alignment.}
    \label{afig:layer_ablation_qual_lie_art}
\end{figure}

For more rigorous evaluation, we conducted a user study with 62 participants. We configured a set with a reference image, a content text prompt, and six synthesized images with different initial noises per method from five competitors. The participants answered below question for 20 sets: Which method best reflects the style in the reference image AND the content in the text prompt?
As indicated in \tref{tab:userstudy}, the majority of participants rated our method as the best.
The lower ratings of the IP-Adapter in the user study may be attributed to its poor text alignment, as illustrated in \fref{fig:vs_competitors_quan} despite its high style similarity.
Examples of the user study are provided in \fref{afig:user_study_ex}.

\begin{table}[h]
\centering
\caption{
\textbf{User study comparison.} We asked participants: Which method best reflects the style in the reference image AND the content in the text prompt?
}
\begin{tabular}{ccccc}
\toprule
Ours & StyleAligned & IP-Adapter & DB-LoRA & StyleDrop \\
\midrule
\textbf{58.15}\% & 13.15\% & 18.47\% & 7.66\% & 2.58\% \\
\bottomrule
\end{tabular}
\label{tab:userstudy}
\end{table}

\newpage

\section{Additional figure results}
\label{sec:add_fig_results}

\begin{figure}[h]
    \centering
    \includegraphics[width=1.0\textwidth]{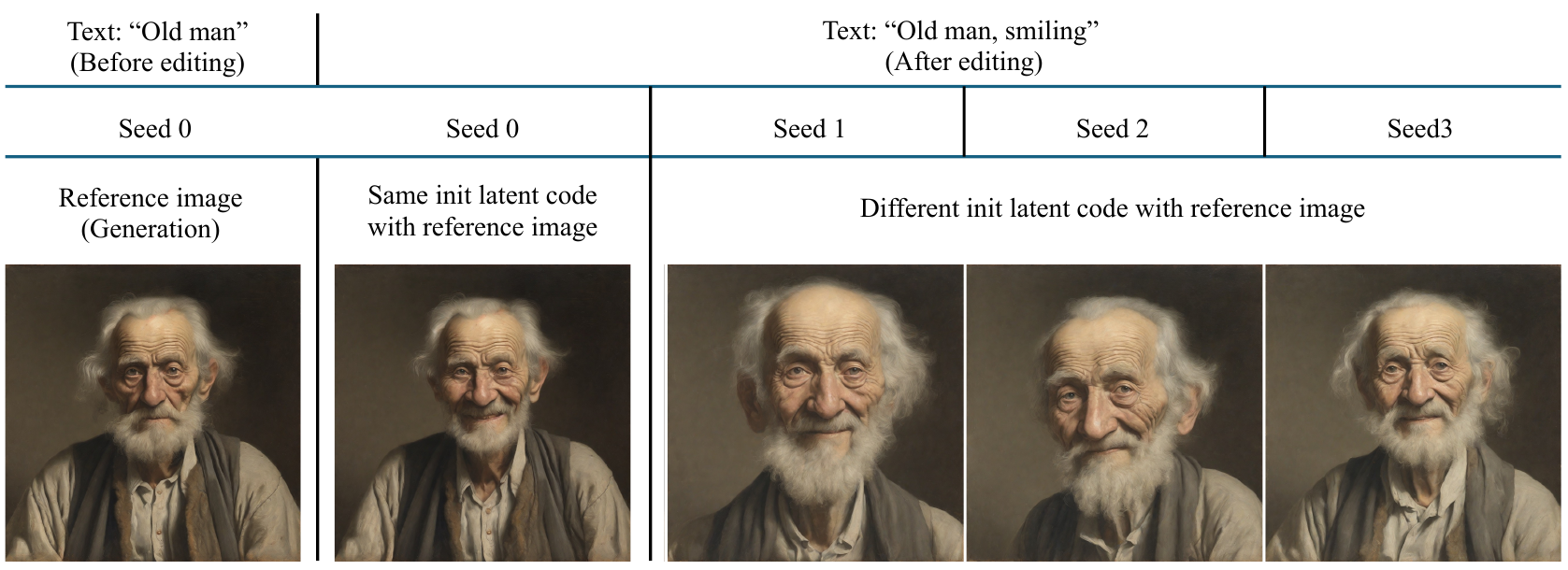}
    \caption{MasaCtrl \cite{cao_2023_masactrl} employ the same initial latent for the reference and inference denoising process for preservation of the target identity. Different initial latent infer the different identities. In contrast, due to the differences in the tasks themselves, ours do not share the same initial latent between the reference denoising process and the inference denoising process.}
    \label{afig:masactrl_different_init_noise}
\end{figure}

\begin{figure}[t]
    \centering
    \includegraphics[width=0.5\linewidth]{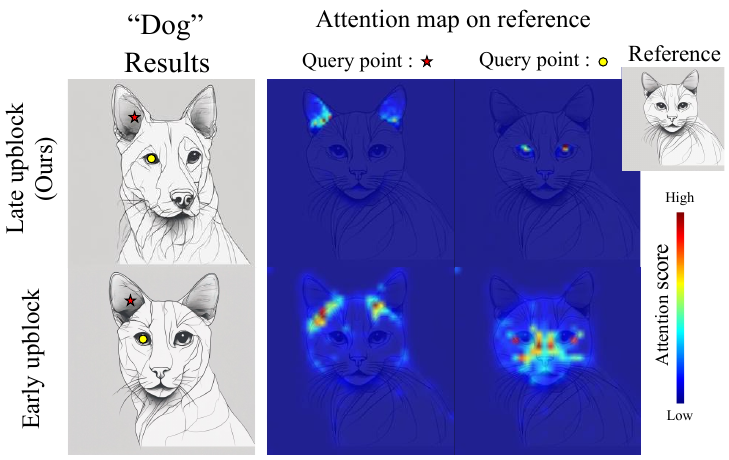}
    \caption{\textbf{Attention map visualization over late and early upblock layers.}
    The late upblock better focuses on the style-corresponding region than the early upblock, leading to more freedom to reassemble small parts. The early upblock attends a larger region leading to content leakage.
    }
    \label{afig:ablation:attentionmap}
\end{figure}

\begin{figure}[h]
    \centering
    \includegraphics[width=0.9\textwidth]{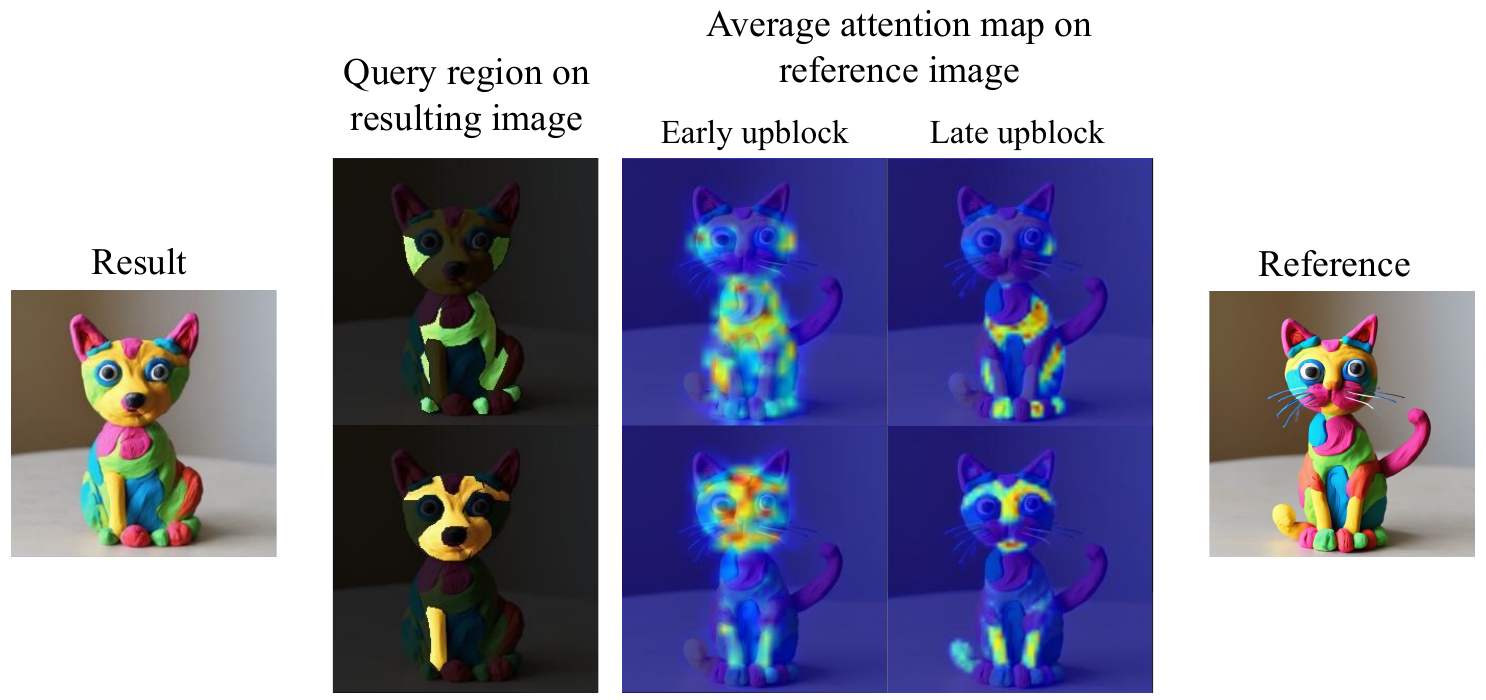}
    \caption{In \fref{fig:ablation:attentionmap}, we only show attention maps of 2 query points. Here, we provide the average attention map of multiple query points on the corresponding query region. At the late upblock, the query point region of the resulting image corresponds to the same style region of the reference image. On the other hand, at the early upblock, the query point region matches not only with the corresponding style region but also with the wider region}
    \label{afig:layer_ablation_visual_advanced}
\end{figure}

\begin{figure}[h]
    \centering
    \includegraphics[width=0.9\textwidth]{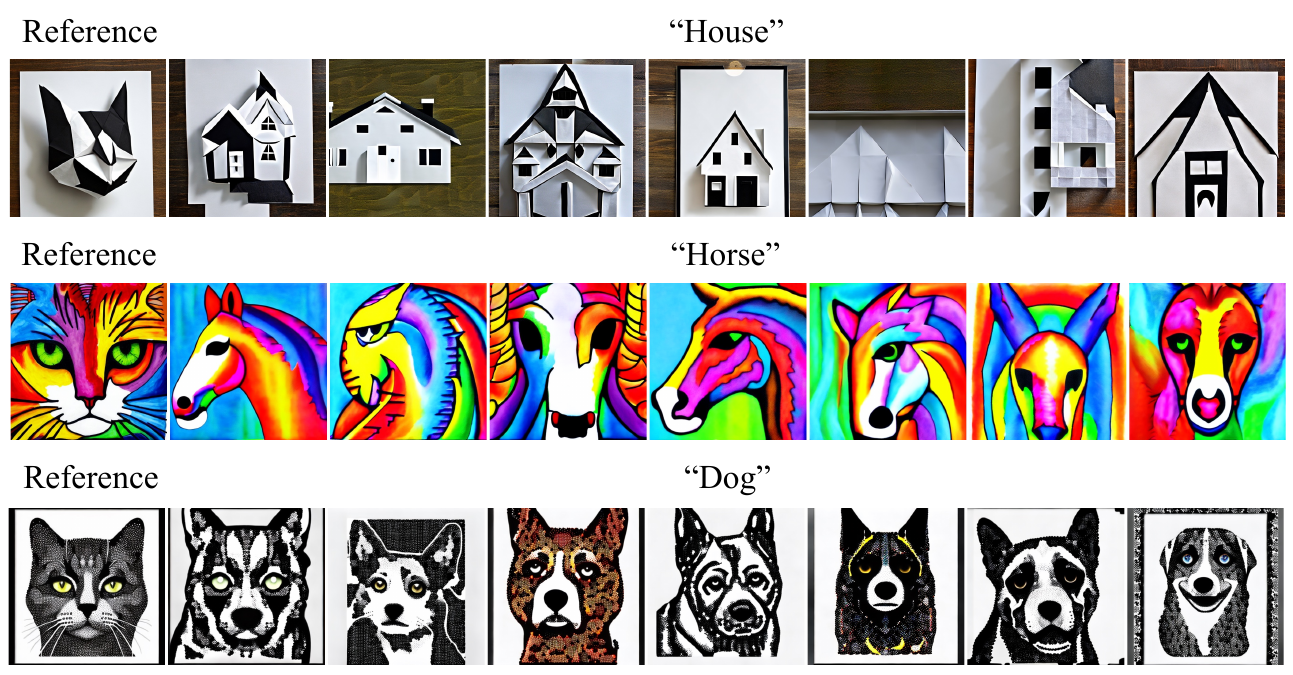}
    \caption{\textbf{Qualitative result of \ours{} on stable diffusion v1.5} Ours also works on the other pretrained diffusion models.}
    \label{afig:SDv15}
\end{figure}

\begin{figure}[h]
    \centering
    \includegraphics[width=0.9\textwidth]{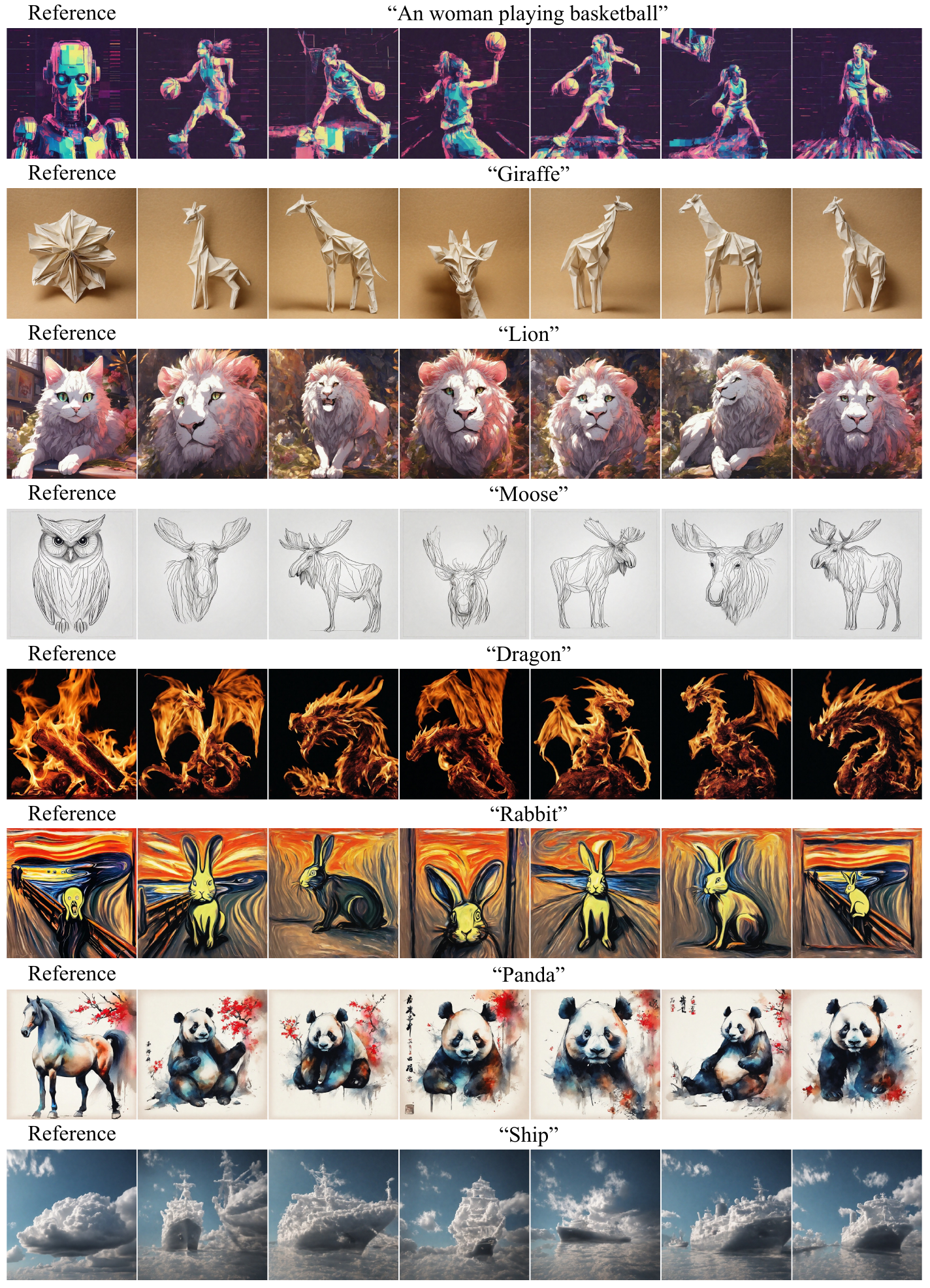}
    \caption{\textbf{Qualitative result of \ours{} within a prompt.} Ours can generate diverse layouts, poses and composition within a prompt.}
    \label{afig:more_diversity_results}
\end{figure}

\begin{figure}[h]
    \centering
    \begin{minipage}[t]{0.49\textwidth}
        \centering
        \includegraphics[width=\textwidth]{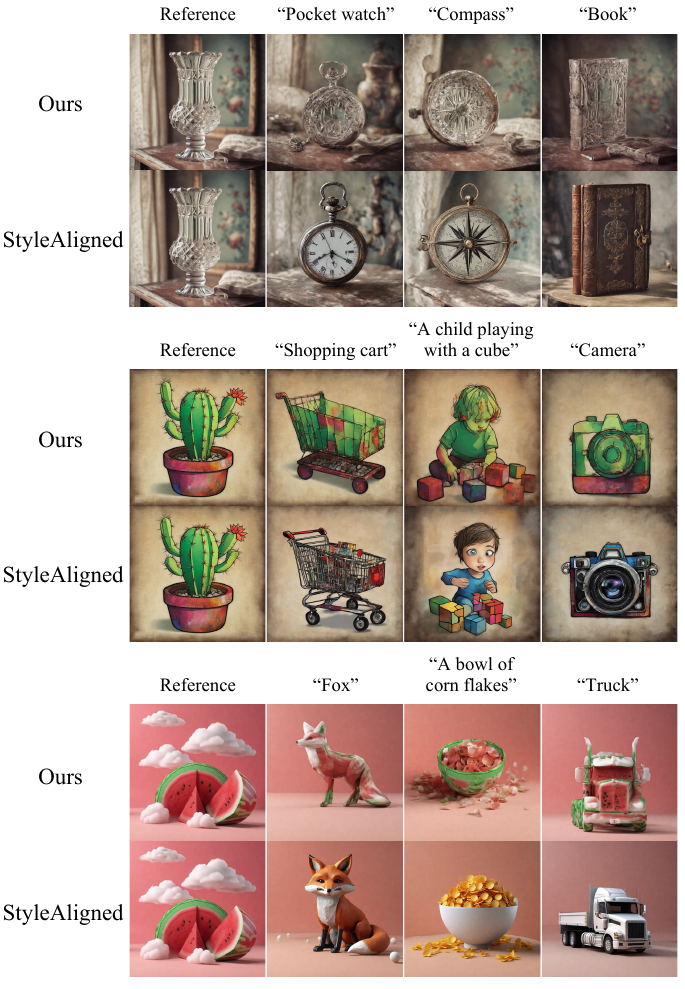}
        \caption{Definition of style is different between ours and StyleAligned.}
        \label{afig:vs_SA_definition_of_style}
    \end{minipage}
    \hfill
    \begin{minipage}[t]{0.49\textwidth}
        \vspace{-200pt}
        \includegraphics[width=\textwidth]{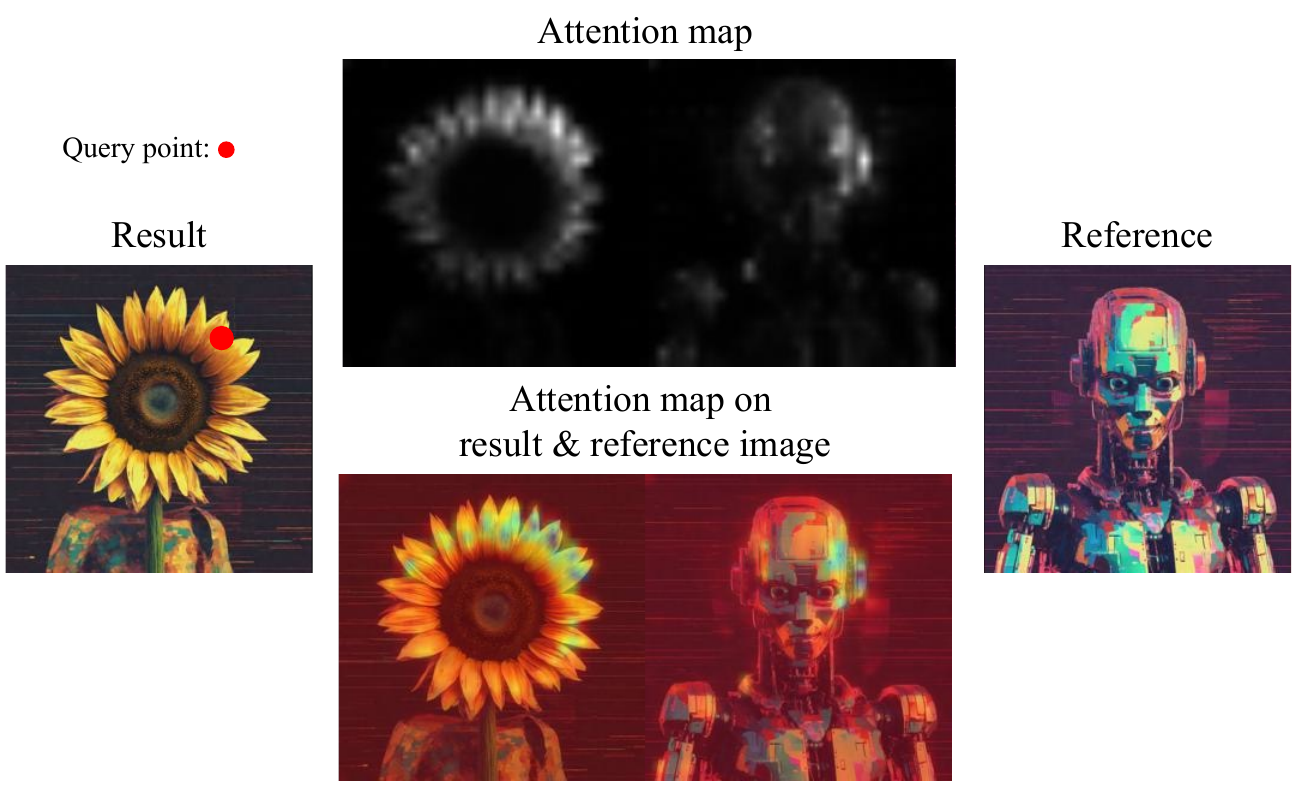}
        \caption{StyleAligned attends both on a reference and a resulting image for shared self-attention mechanism. In contrast, \ours{} only attends on a reference features which leads to better reflection of style in the reference image.}
        \label{afig:vs_SA_attn}
    \end{minipage}
\end{figure}

\begin{figure}[h]
    \centering
    \includegraphics[width=0.8\linewidth]{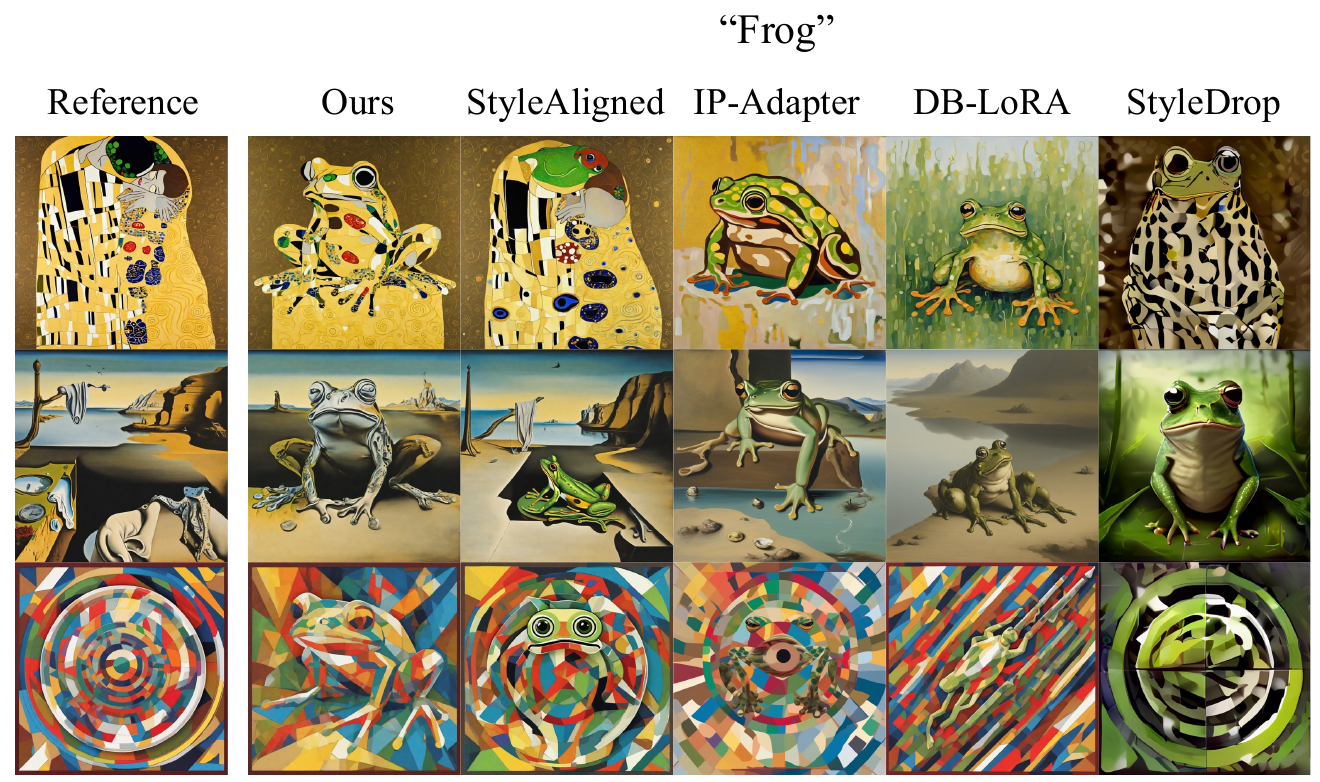}
    \caption{\textbf{Qualitative comparison with varying styles and fixed content.} \Ours{} reflects style elements from various reference images to render ``frog" while others struggle.}
    \label{fig:vs_competitor_frog}
\end{figure}



\begin{figure}[t]
    \centering
    \includegraphics[width=0.9\linewidth]{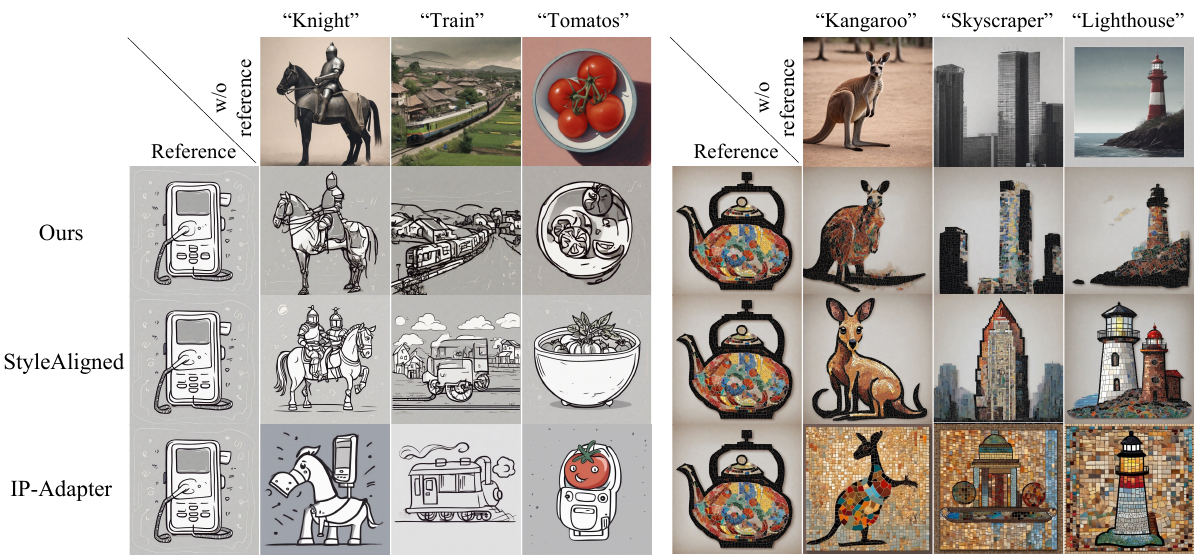}
    \caption{\textbf{Comparison of contents change while reflecting the style in the reference.} Each column shares the same initial noise. \Ours{} reflects the style in the reference with minimal changes in the contents of the original denoising process while others produce drastic changes. } 
    \label{afig:ablation:init_noise_is_content}
\end{figure}

\begin{figure}[h]
    \centering
    \includegraphics[width=1.0\linewidth]{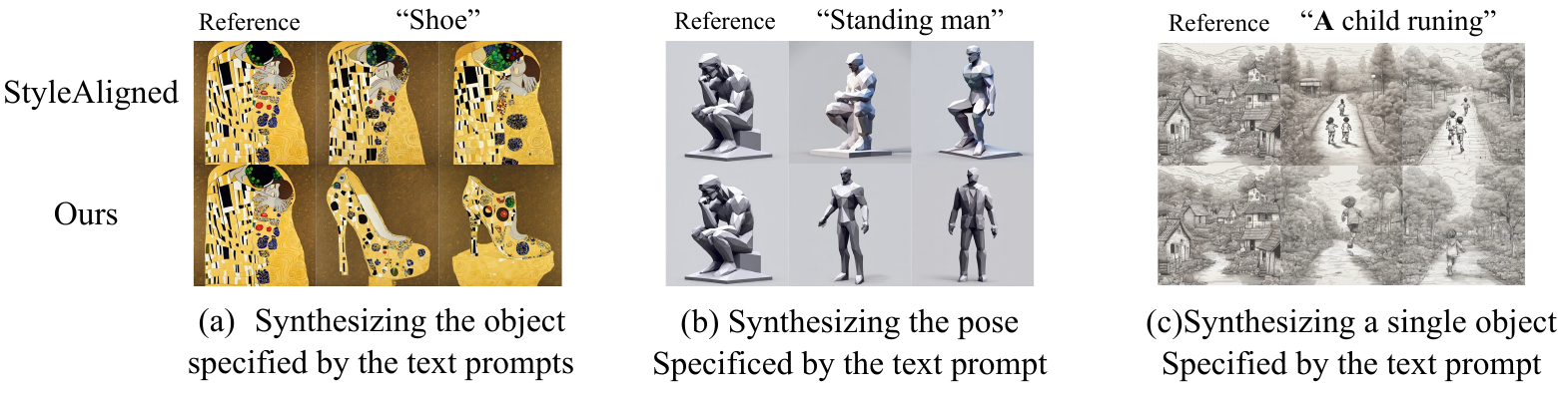}
    \caption{    \textbf{Comparison for content leakage.} While StyleAligned suffers from content leakage from the reference, results from ours clearly align with the text prompts}
    \label{afig:ablation:vs_SA_senario}
\end{figure}

\begin{figure}[h]
    \centering
    \includegraphics[width=1.0\linewidth]{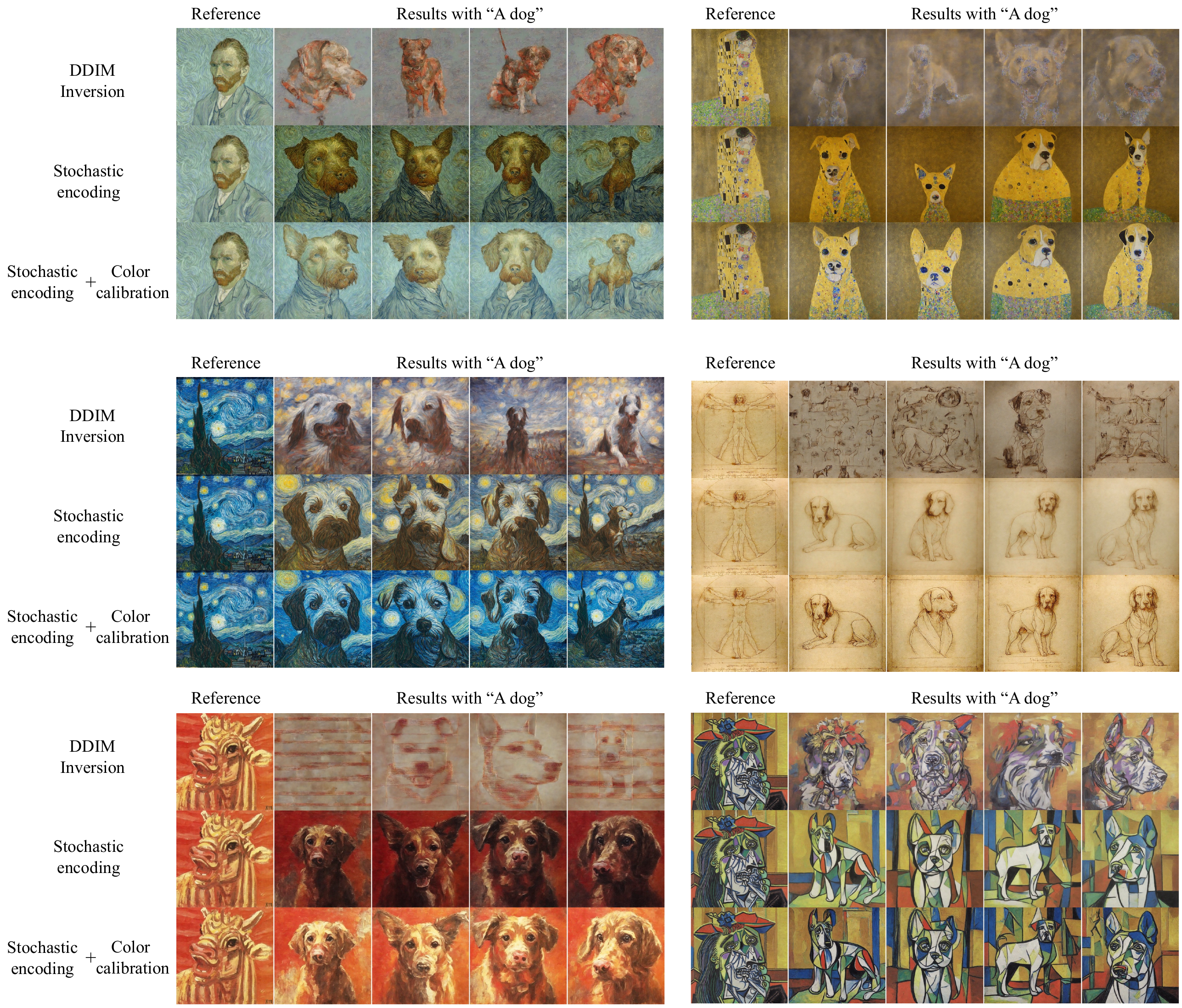}
    \caption{More comparison of DDIM inversion with stochastic encoding, and with stochastic encoding and color calibration. Stochastic encoding reduces artifacts in the generated images, while color calibration better preserves the colors of the reference image.}
    \label{afig:vsp_add_noise_calibration_six_exp}
\end{figure}

\begin{figure}[h]
    \centering
    \includegraphics[width=0.8\linewidth]{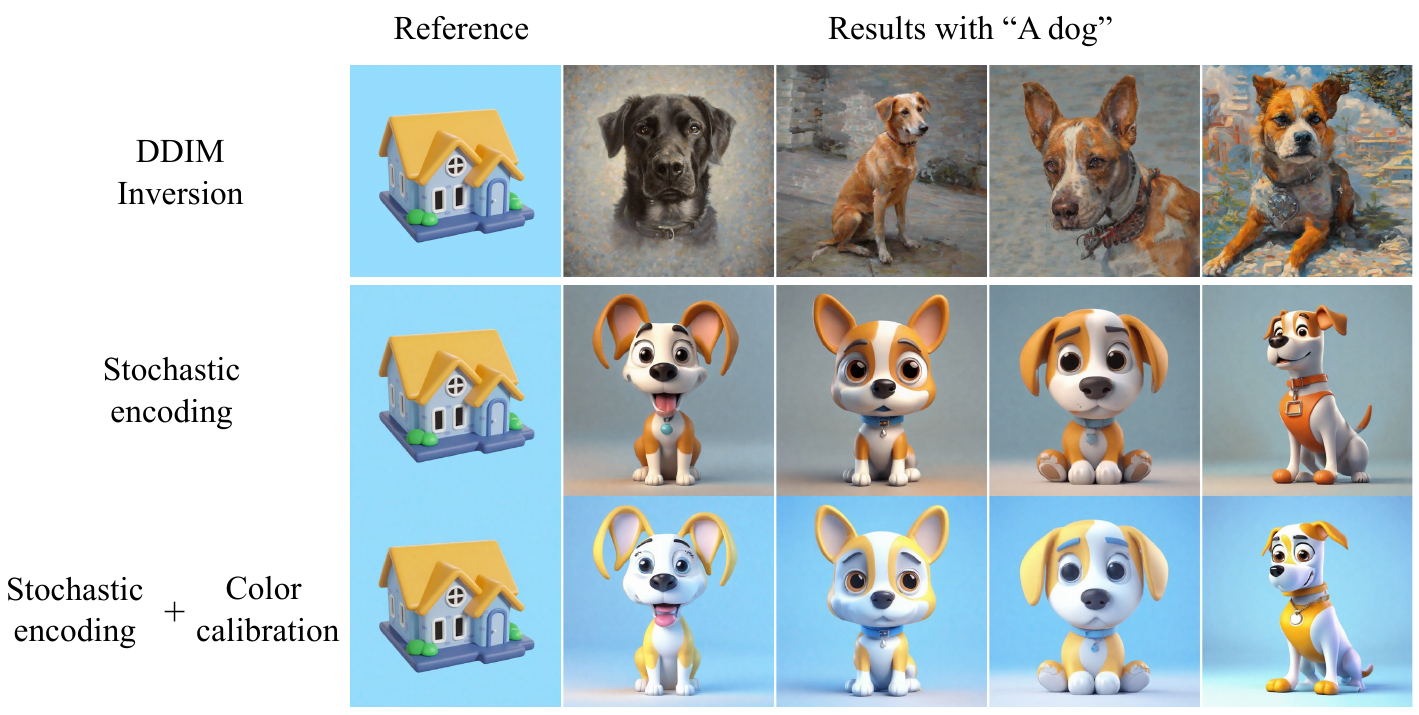}
    \caption{\textbf{Ablation study with StyleAligned \cite{hertz2023style}.} Our strategy for a real reference image is compatible with self-attention variants for  boosting the reflection of style elements.}
    \label{afig:ablation:add_noise_with_SA}
\end{figure}

\begin{figure}[h]
    \centering
    \includegraphics[width=0.8\linewidth]{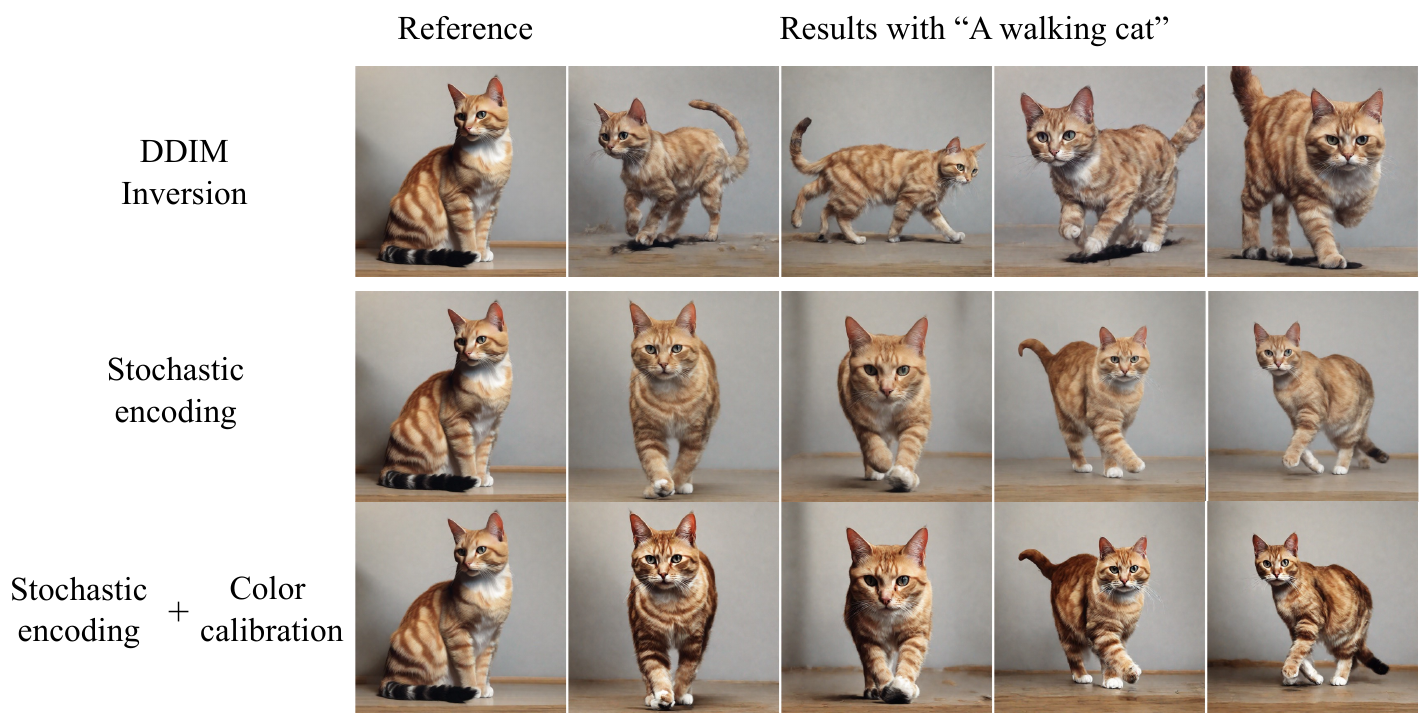}
    \caption{\textbf{Ablation study with MasaCtrl \cite{cao_2023_masactrl}.} Our strategy for a real reference image is compatible with self-attention variants for reducing artifacts and boosting the reflection of visual elements.}
    \label{afig:ablation:add_noise_with_masactrl}
\end{figure}

\begin{figure}[h]
    \centering
    \includegraphics[width=0.8\linewidth]{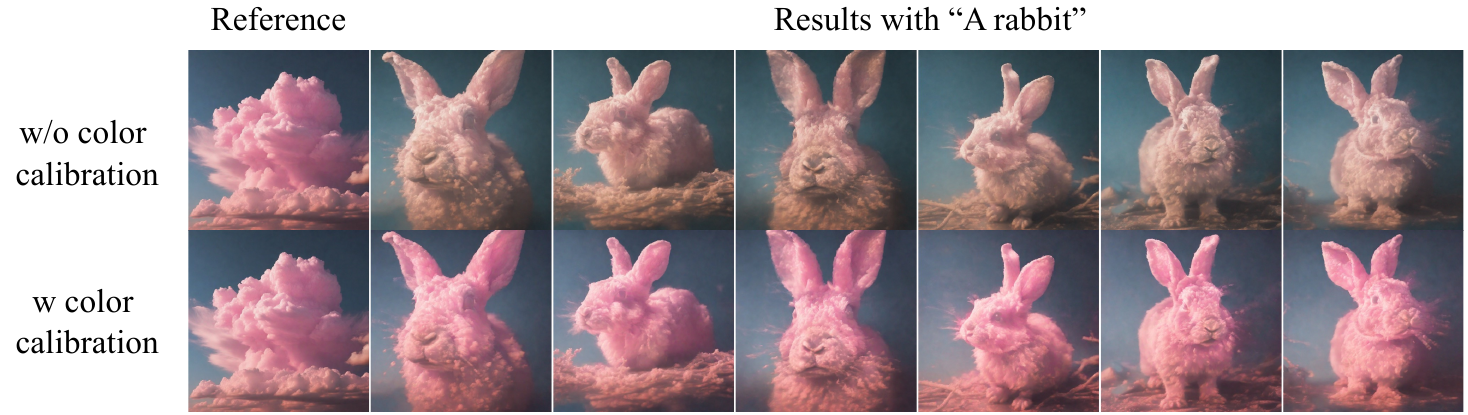}
    \caption{\textbf{Color calibration ablation study with generating process from random noise.}. Color calibration also decreases the minor discrepancy of color between the reference image and the resulting image that barely occurs in the generation setting.}
    \label{afig:color_calibration_vsp}
\end{figure}

\begin{figure}[h]
    \centering
    \includegraphics[width=1.0\linewidth]{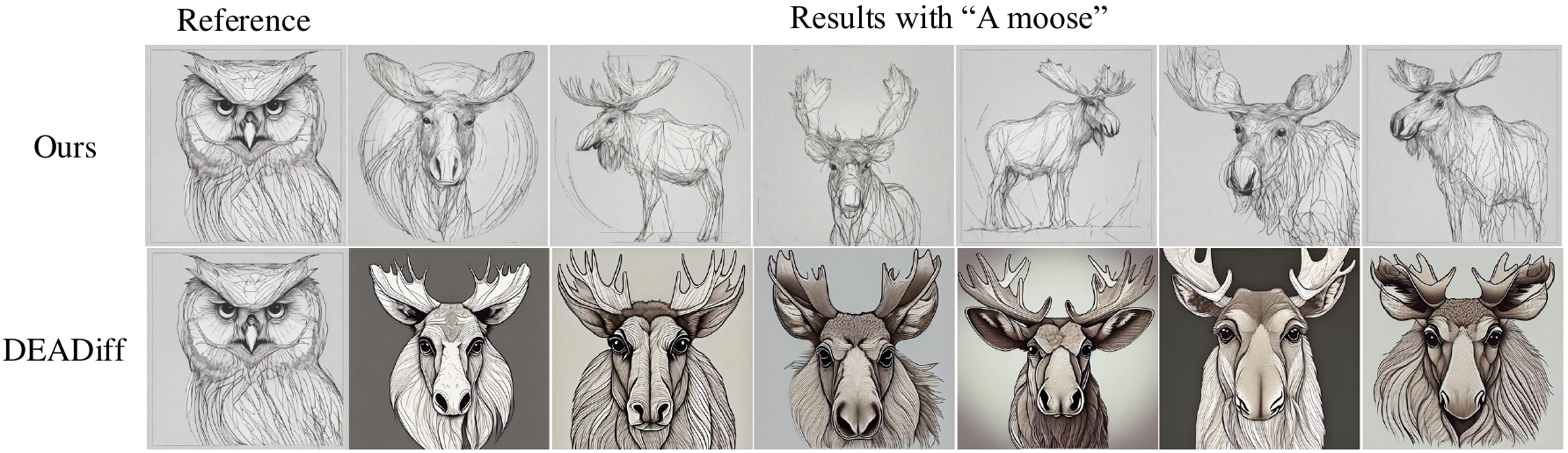}
    \caption{\textbf{Content leakage in DEADiff \cite{qi2024deadiff}} The results of DEADiff suffer from content leakage (e.g., the frontal face of “A moose” and the “An owl” reference image), which reduces diversity within a text prompt. \Ours{} do not suffer content leakage while reflecting style.}
    \label{afig:content_leakage_deadiff}
\end{figure}

\begin{figure}[h]
    \centering
    \includegraphics[width=1.0\linewidth]{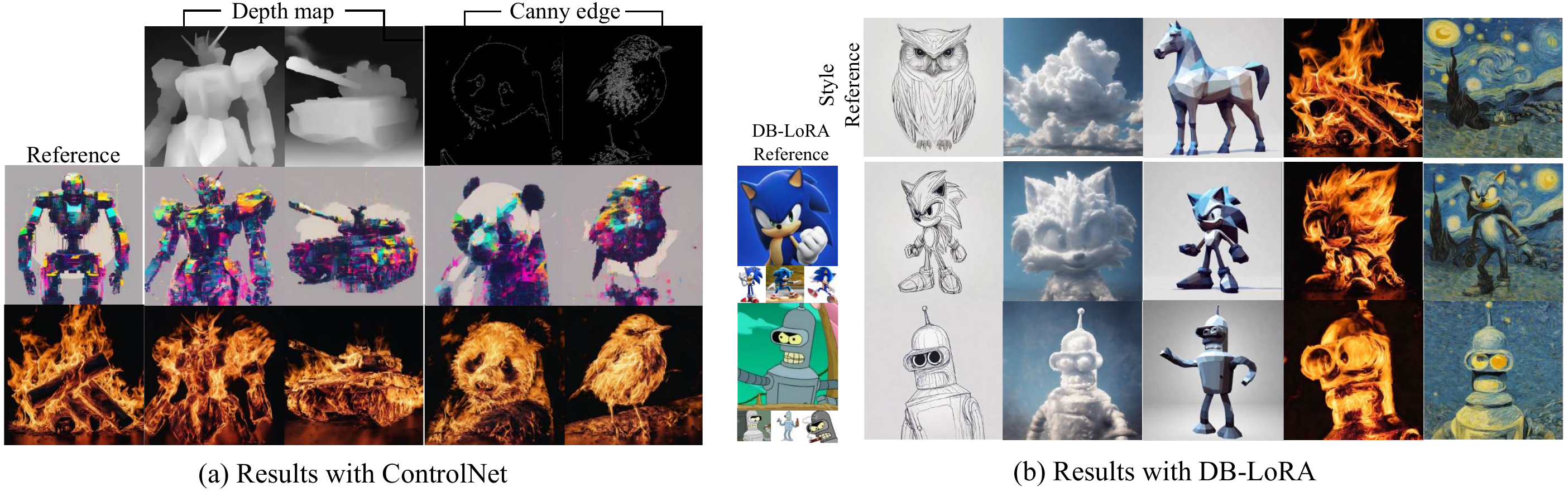}
    \caption{\textbf{Visual style prompting with existing techniques.} Our method is compatible with ControlNet and Dreambooth-LoRA.}
    \label{fig:controlnet_db_lora}
\end{figure}

\begin{figure}[h]
    \centering
    \includegraphics[width=0.9\linewidth]{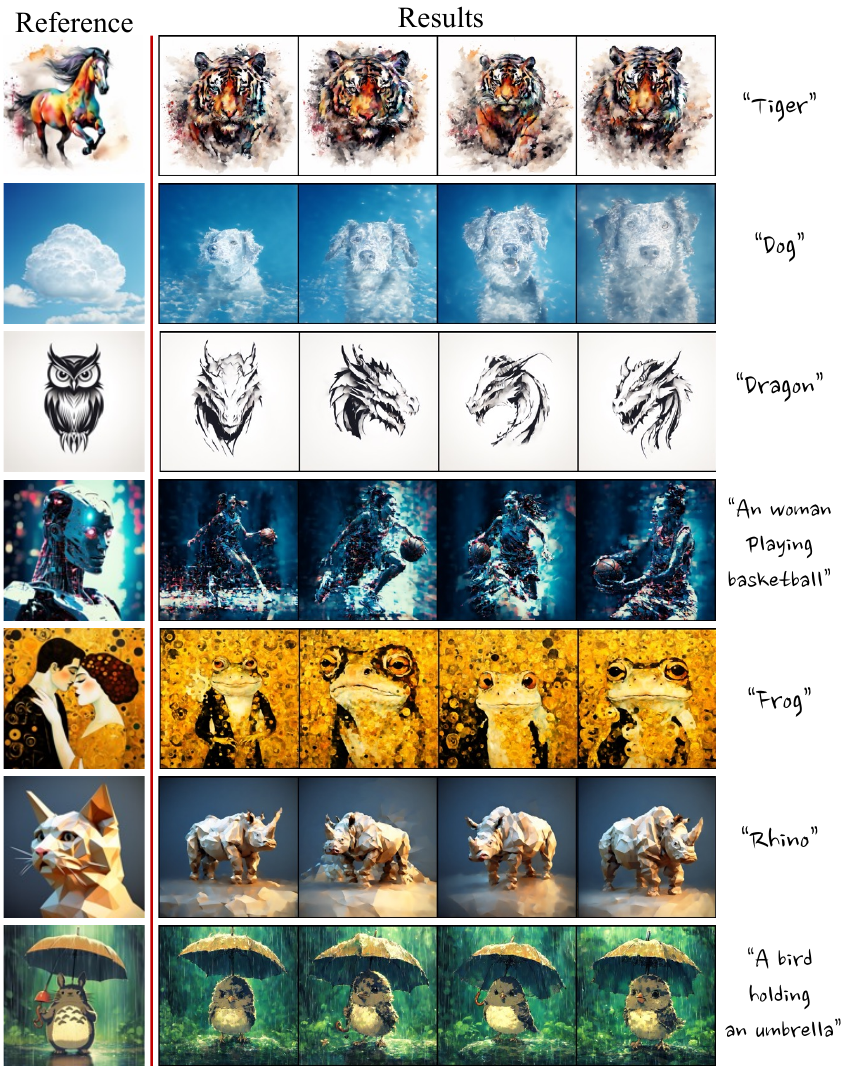}
    \caption{Qualitative result of visual style prompting in Pixart-$\alpha$.}
    \label{fig:pixart}
\end{figure}

\begin{figure}[h]
    \centering
    \includegraphics[width=1.0\linewidth]{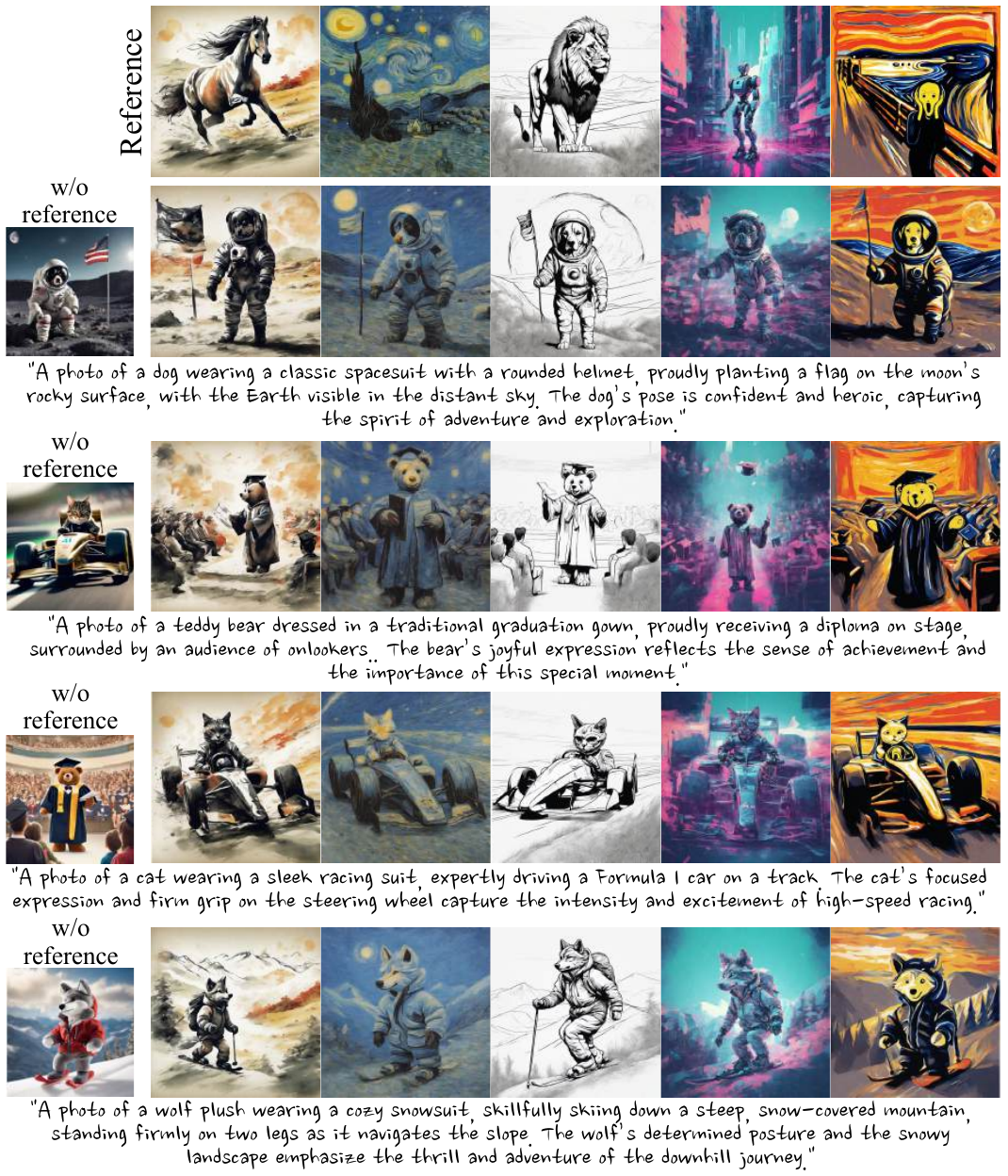}
    \caption{Qualitative result of visual style prompting in complex text prompts.}
    \label{fig:complex_prompt}
\end{figure}

\begin{figure}[h]
    \centering
    \includegraphics[width=1.0\linewidth]{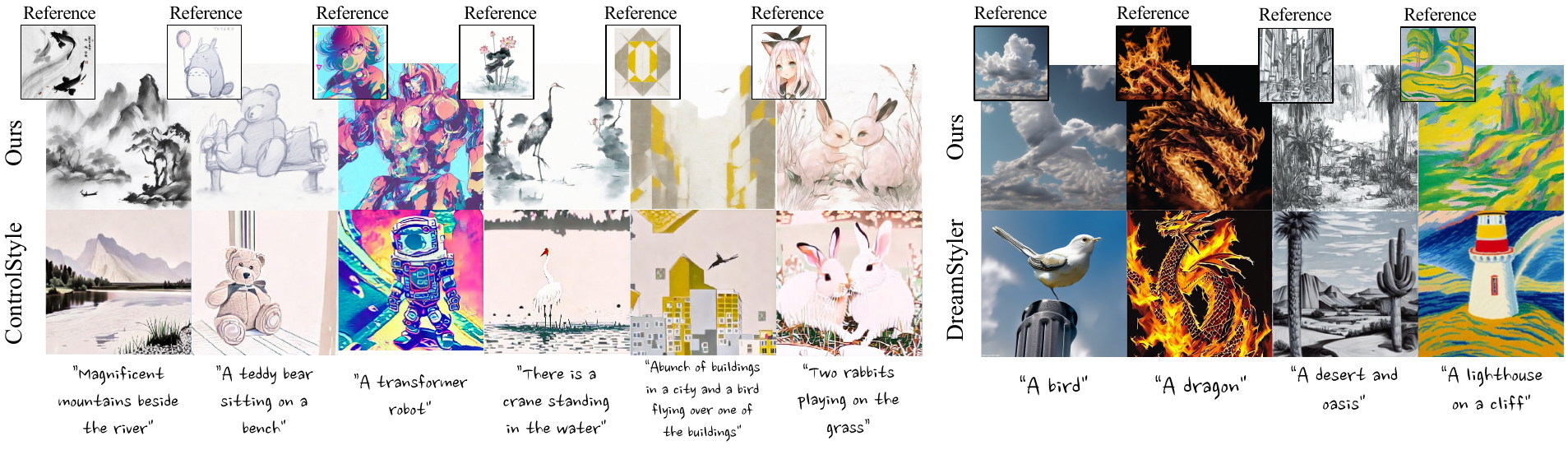}
    \caption{Comparison of ours with DreamStyler and ControlStyle.}
    \label{fig:compare_dream}
\end{figure}

\begin{figure}[h]
    \centering
    \includegraphics[width=1.0\linewidth]{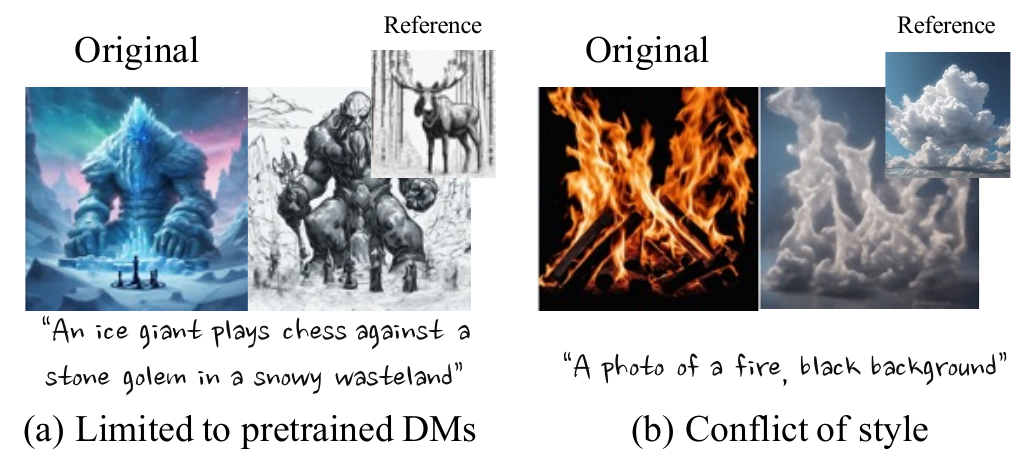}
    \caption{\textbf{Limitation: }\textbf{(1) Text Alignment Failure}: Missing 'stone golem' from the prompt (a), and \textbf{(2) Prompt Conflict}: 'Cloud in the sky' vs. 'fire, black background' (b).}
    \label{fig:limit}
\end{figure}





\end{document}